\documentclass[draft,onecolumn]{IEEEtran}

\pdfoutput=1

\usepackage{ieeefig,graphicx,amssymb,amsmath,cite,algorithm,algorithmic}

\begin{document}

\title{Pattern Encoding on the Poincar\'{e} Sphere}

\author{Aleksandra Pi\v{z}urica
\thanks{A. Pi\v{z}urica is with the Department
Telecommunications and Information Processing, IPI-TELIN-iMinds, Ghent University, Belgium\protect\\
E-mail: aleksandra.pizurica@telin.ugent.be
}}

\markboth{DRAFT}%
{Pi\v{z}urica: Pattern encoding on the Poincar\'{e} sphere}

\maketitle

\begin{abstract}
This paper presents a convenient graphical tool for encoding visual patterns
(such as image patches and image atoms) as point constellations in
a space spanned by perceptual features and with a clear
geometrical interpretation. General theory and a practical pattern encoding scheme are presented, inspired by encoding polarization
states of a light wave on the Poincar\'{e} sphere. This new pattern
encoding scheme can be useful for many applications in image
processing and computer vision. Here, three possible applications
are illustrated, in clustering perceptually similar patterns,
visualizing properties of learned dictionaries of image atoms and
generating new dictionaries of image atoms from spherical codes.
\end{abstract}






\section{Introduction}\label{Sec:introduction}

\begin{figure}[t]
\begin{center}
\includegraphics[width=0.55\textwidth]{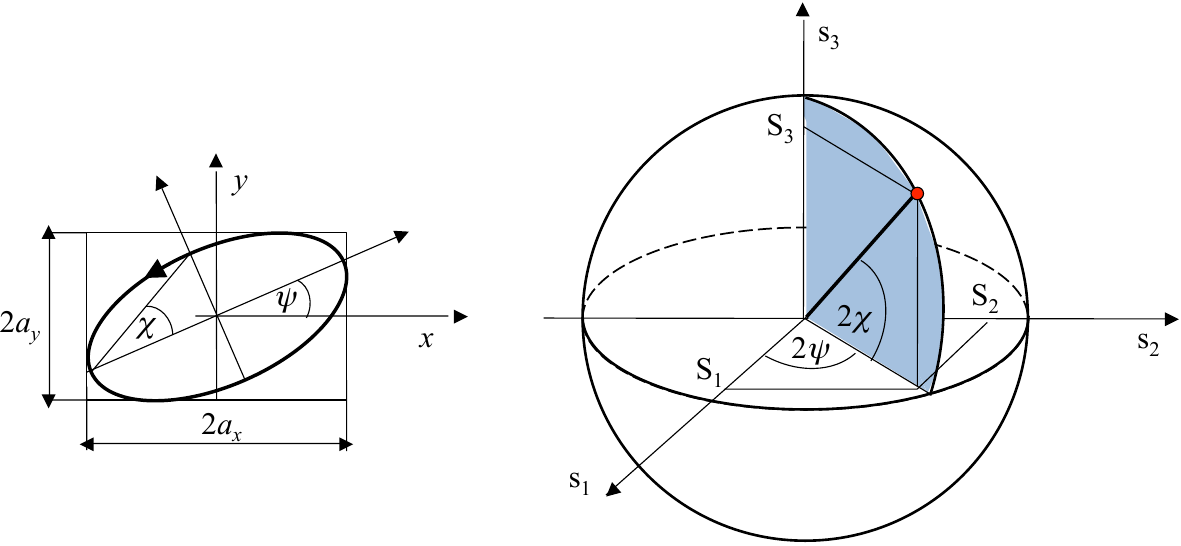}
\caption{Representation of light polarizations on the Poincar\'{e} sphere. {\bf Left:} Three parameters of the polarization ellipse:
direction angle $\psi$,
ellipticity angle $\chi$ and orientation (for the opposite orientation, the ellipticity angle is $-\chi$). {\bf Right:} The corresponding representation with Stokes parameters \cite{BornWolf} on the Poincar\'{e} sphere. \label{Fig:PolarizationEllipseSphere}} 
\end{center}
\end{figure}

\begin{figure}[t]
\begin{center}
\begin{tabular}{cc}
\includegraphics[width=0.6\textwidth]{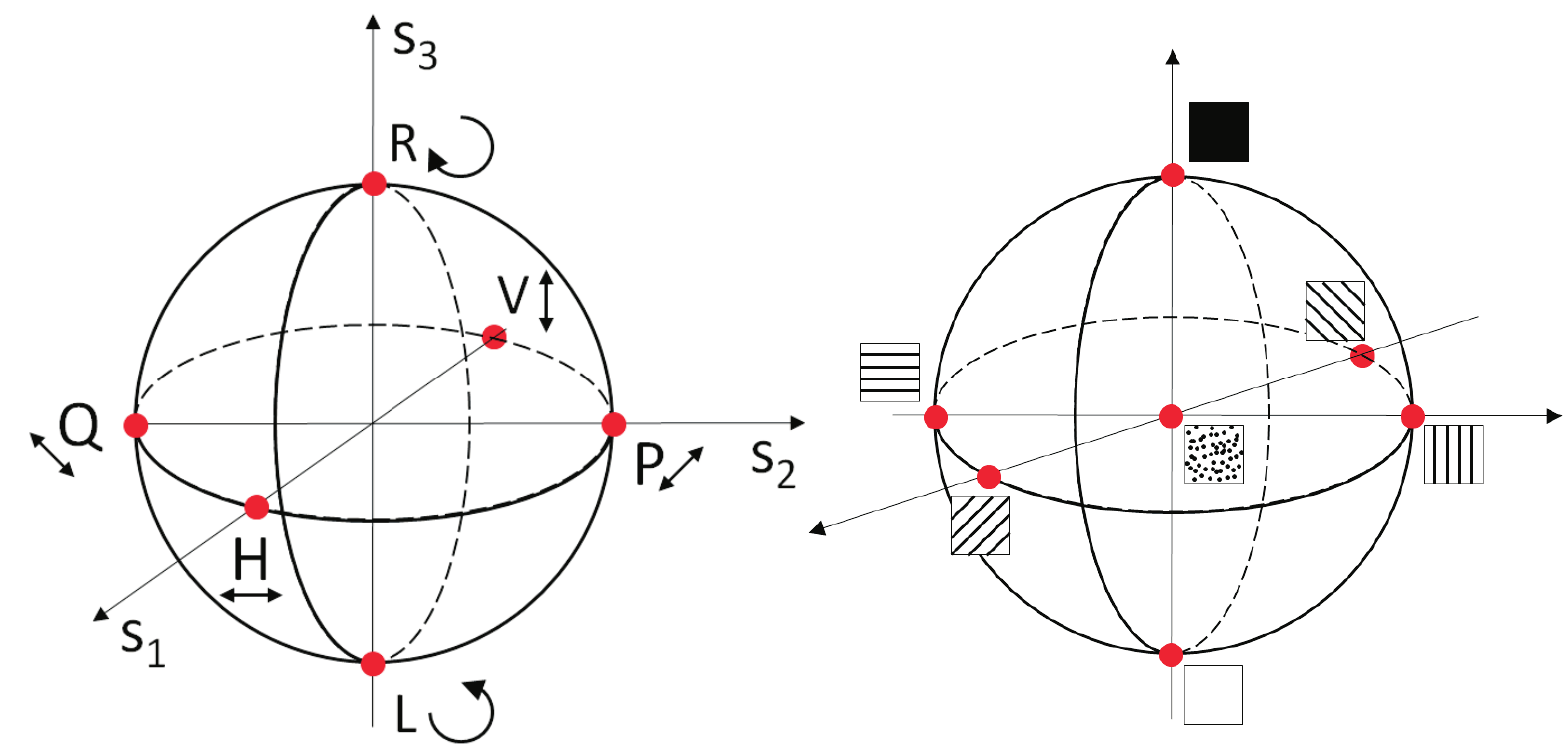}
\end{tabular}
\caption{Pattern encoding on the Poincar\'{e} sphere: a concept showing 
encoding of polarization states and an analogy for
visual patterns. \label{Fig:PoincarePolarizations}}
\end{center}
\end{figure}

Many remarkable developments in computer vision and machine
learning were initiated by analogies with physical systems and by
clear geometrical interpretations. This work on pattern encoding
is inspired by a beautiful geometrical representation of light
polarizations on the Poincar\'{e} sphere \cite{BornWolf,Malykin97} 
illustrated in Fig.~\ref{Fig:PolarizationEllipseSphere}. In this
representation, each polarization state of a fully polarized light
wave corresponds to one point on the surface of a unit sphere\footnote{For a planar monochromatic wave, the trajectory of the electric field vector with components 
$E_x =a_x\cos(\tau+\phi_x)$, $E_y =a_y\cos(\tau+\phi_y)$ and $E_z=0$ in the $z$-direction (perpendicular to the wave propagation) is elliptical: 
$\Bigl(\frac{E_x}{a_x}\Bigr)^2 + \Bigl(\frac{E_y}{a_y}\Bigr)^2 -2\frac{E_x}{a_x}\frac{E_y}{a_y}\cos\delta = (\sin\delta)^2$, with $\delta=\phi_x-\phi_y$ (see Fig.~\ref{Fig:PolarizationEllipseSphere}). Each polarization state, characterized by the direction $\psi$,
ellipticity $\chi$ and orientation of the polarization ellipse, or alternatively, by the three Stokes parameters $S_1=a^2_x-a^2_y$, $S_2=2a_xa_y\cos\delta$, $S_3=2a_xa_y\sin\delta$, can be presented as a point with coordinates: $S_1=S_0\cos(2\chi)\cos(2\psi)$, $S_2=S_0\cos(2\chi)\sin(2\psi)$, $S_3=S_0\sin(2\chi)$, where the fourth Stokes parameter is $S_0=a^2_x+a^2_y$. For partially polarized light,  $S_0^2 \leq S^2_1+S^2_2+S^2_3$ and $\sqrt{(S^2_1+S^2_2+S^2_3)}/S_0$ is the degree of polarization. For further details, see \cite{BornWolf}. In normalized coordinates (after division by $S_0$) the points are on a unit sphere.}. All
linear polarizations are on the equator and the ellipticity
increases towards the poles, where it becomes circular (see
Fig.~\ref{Fig:PoincarePolarizations}, left). The northern and the southern hemisphere correspond each to one orientation of the ellipse. Partially polarized light can be encoded with a sequence of concentric spheres, each corresponding to a particular degree of polarization. Moving away from the
surface of the unit sphere inwards, the degree of polarization
diminishes, reducing to zero in the origin. Some readers may be familiar with this representation from digital
communications too: in optical communications with POLarization
Shift Keying (POLSK) modulation \cite{Benedetto92,Yu13,Borkowski14,Pizurica98}, symbols
to be transmitted (i.e., codewords) are polarization states
represented as points on the Poincar\'{e} sphere.

Here we introduce an idea for similar encoding of visual patterns
and demonstrate possible benefits of such a representation. A
conceptual scheme is illustrated in
Fig.~\ref{Fig:PoincarePolarizations}(b) where the degree of
regularity is seen as the degree of orientedness and treated
equivalently to the degree of ``polarization''. Fully oriented
patterns are on the surface of the unit sphere and the regularity
decreases (i.e., randomness increases) in the direction of the
origin, where the pattern turns to random noise. Note that in the same way as we encode partially polarized light on a ball rather than on a sphere, we also represent visual patterns within a unit ball, where each shell (each of the concentric spheres) within this ball corresponds to a particular regularity level. With slight abuse of notation, we refer to encoding on the sphere when no confusion is possible. The
two opposite poles correspond to two colorings (``black'' and
``white'') and a particular elevation corresponds to a given
balance between dark and bright (e.g., to mean patch
intensity).

This type of pattern representation can be useful for 
non-local image processing, such as non-local
image denoising \cite{Buades05, Dabov07, Kervrann08, Mairal09, Katkovnik10, Chatterjee12, Milanfar13}, patch-based inpainting
\cite{Criminisi07,Guillemot14} and other problems where it is of interest
to accelerate the search for well matching patches or surroundings, as well as in
image content analysis and visualizing some properties of learned
dictionaries of image atoms, as we demonstrate later on.

Arrangements of points on spheres have been a subject
of great interest in mathematics, physics and engineering, from
Kepler and Gauss to date \cite{Sloane81,Hardin96,Lazic87,Hopkins10}, as well as subspace packings, or more general arrangements of $n$-dimensional subspaces of $\mathbb{R}^m$ \cite{Calderbank99}. For a comprehensive review of spherical
codes (optimal, according to a given criterion, constellations of
points on a sphere) see \cite{ConwaySloaneBook}. We will show
later how these spherical codes can be translated into
dictionaries of image atoms, using the developed pattern encoding
scheme.


In our construction, we need to make some particular choices in
terms of features that will correspond to the three available
dimensions, either after dimensionality reduction applied to some ``bag of features'' or by selecting some obvious perceptual dimensions. Already in the
pioneering studies on visual pattern discrimination by Bela Julesz
\cite{Julesz62} it has been pointed out that ``periodic structures
attract the eye'' and that visual system tries to ``connect'' the
bright values on the one hand and the dark ones on the other,
perceiving them as separate clusters. Classical studies on 
texture perception \cite{Francos93,Liu96}, often put forward
three dimensions as the most important ones, corresponding either
to \emph{repetitiveness}, \emph{directionality} and
\emph{granularity and complexity} or to the three Wold features:
\emph{periodicity}, \emph{directionality} and \emph{randomness}.
The overall brightness of an image pattern (or image patch) is one
of the most discriminant properties too (even though in many
applications noticing differences in lighting might not be of
interest and invariance to lighting might be an asset). Many other
useful features can be derived from the co-occurrence matrices
of Haralick \cite{Haralick79} and from
histograms in different subbands of a filter bank decomposition
\cite{Unser95,Portilla00,Do02,Lasmar14}. Much work has been done
on texture characterization using Markov Random Field models
\cite{Cross83, Chelappa85, Derin87, Krishnamachari97}, and
structural models \cite{Ahuja81, Ojala96}. It is not our intention
here to give an exhaustive overview of texture analysis
literature. The interested reader should consult some of the
excellent textbooks and reviews on this topic, such as
\cite{Prat78,Petrou06}. We attempt here to
capture only some projections of a ``galaxy'' of textural
features, as they were named in \cite{Xie08}, onto lower
dimensional spaces with nice geometrical structure and
interpretation, providing an alternative or a complimentary tool to well established methods for extraction and visualization of textural features, based on the principal component analysis \cite{Kim01} and multidimensional scaling \cite{Beatty97}.

In this paper, we construct only one specific set of parameters
based on which we build our graphical tool that demonstrates
the potentials of the proposed approach. We shall discuss some
possibilities for generalizations that will be studied in
follow-up papers. The examples that will be presented in this
paper serve only to prove the concept, while further work is
needed to actually employ the proposed technique in the
actual computer vision
applications.

The paper is organized as follows. Section \ref{Sec:GeneralConcept} introduces a general theory or concept of visual pattern encoding on the Poincar\'{e} sphere and some possible generalizations. While the proposed pattern encoding scheme is three dimensional, we also discuss possible extensions to a fourth dimension, which encodes the scale explicitly. In Section \ref{Sec:ToyEncodingScheme}, we develop a practical pattern encoding tool based on the following three features: dominant orientation, regularity and mean intensity. While the developed scheme allows one to use any available method for estimating dominant pattern orientation and its regularity, we introduce here also some low-complexity methods that can serve this purpose. The proposed method for dominant orientation estimation is applicable even to very small image patches and works reasonably well in the presence of noise and blur. We show how the developed graphical tool maps conveniently image patches with different dominant orientations and different regularity levels to the corresponding point constellations. Three possible applications are demonstrated in Section \ref{Sec:Applications}: patch clustering, visualizing the properties of learned dictionaries of image atoms and generating dictionaries of image atoms from spherical codes. Concluding remarks are in Section \ref{Sec:Conclusion}.

\begin{figure}[t]
\begin{center}
\includegraphics[width=0.3\textwidth]{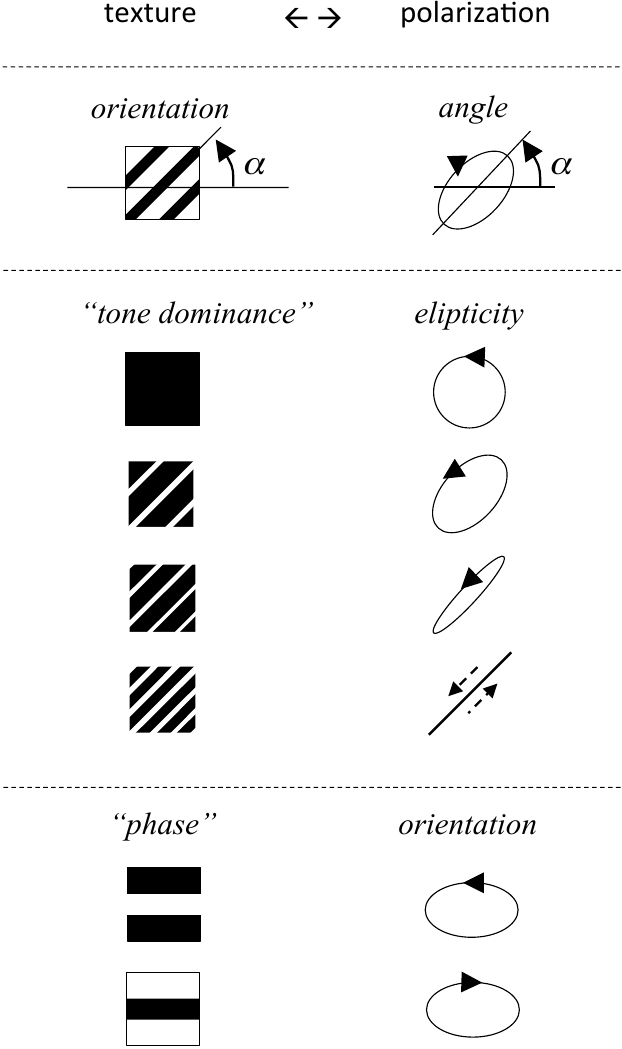}
\caption{Some possible parallels between visual pattern features and polarization parameters.\label{Fig:PolarizPatternAnalogy}}
\end{center}
\end{figure}

\section{A general concept} \label{Sec:GeneralConcept}

The aim of this work is to construct a simple and elegant graphical tool for encoding visual patterns. Extensions and generalizations are likely to follow later on, since (quoting Henri Poincar\'{e}) ``it is the analogy with geometry which can create fruitful associations of ideas and suggest useful generalizations'' \cite{Poincare1895}.

We can think of many interesting parallels between characterizing
light polarizations and visual patterns.
A fully polarized light wave is represented by three parameters
\cite{BornWolf}: the \emph{angle}, \emph{elipticity} and
\emph{orientation} of the polarization ellipse (Fig.~\ref{Fig:PolarizationEllipseSphere}). In texture
perception studies, three most important perceptual dimensions are
often described as \emph{periodicity},
\emph{directionality} and \emph{randomness} \cite{Francos93,Liu96}. The degree of
polarization can be translated into the degree of pattern regularity. With this
dimension, we can travel in both cases from the origin (that encodes fully non-polarized, i.e., completely irregular, random
patterns) towards the surface of the sphere where we have fully
polarized, i.e., completely regular patterns\footnote{Suppose here that we have a particular definition of
pattern regularity, which can be user-defined or customized.}. An obvious analogy to the angle of the polarization
ellipse is the angle of dominant orientation in a visual
pattern. The meaning of ``orientation'' in the case of
polarization is binary (traversing the ellipse 
clockwise or counter-clockwise). A possible analogy for greyscale patterns is the dominance of bright or dark tones, each corresponding to one hemisphere. Finally, the ellipticity of the polarization can be seen as analogous to the level of dis-balance between dark and bright tones (converging to the two opposite extremes at the poles). 
Fig.~\ref{Fig:PolarizPatternAnalogy} illustrates these analogies on fully oriented binary patterns.

\begin{figure}[t]
\begin{center}
\begin{tabular}{cc}
\includegraphics[width=0.6\textwidth]{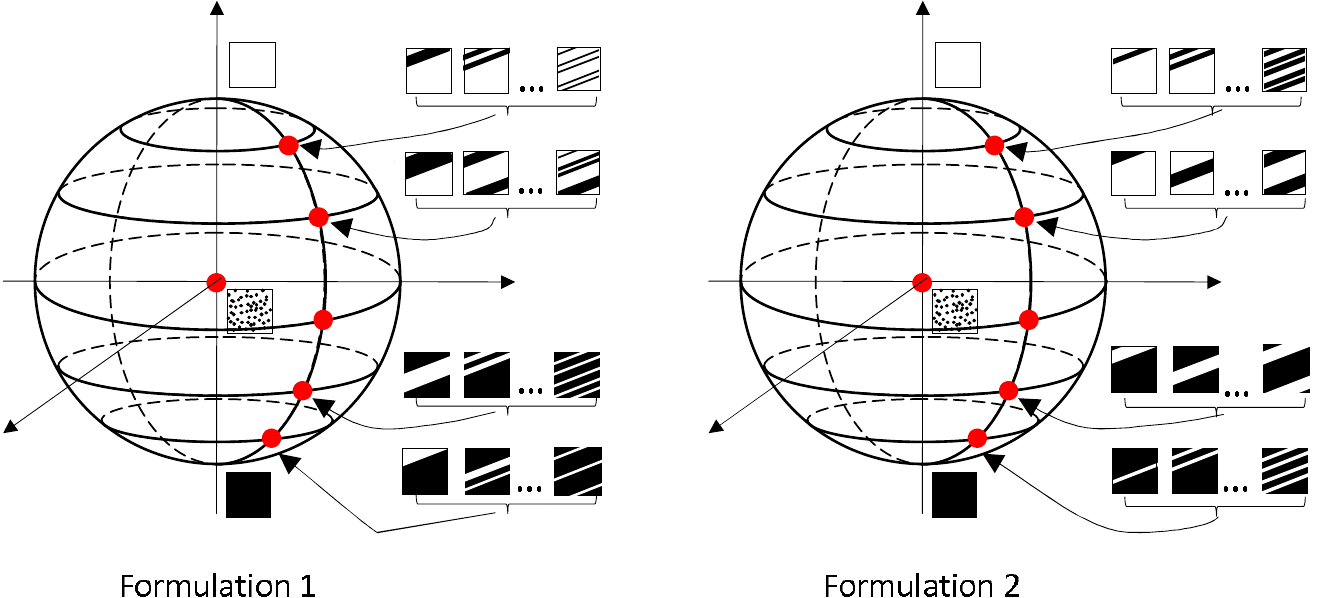}
\end{tabular}
\caption{Two pattern encoding schemes where the elevation angle corresponds to two different formulations of the dark-bright balance: mean patch intensity (formulation 1, left) and a granularity level or the maximum width of lines or blobs (formulation 2, right).
\label{Fig:ElevationFormulations}}
\end{center}
\end{figure}

\begin{figure}[t]
\begin{center}
\begin{tabular}{cc}
\includegraphics[width=0.4\textwidth]{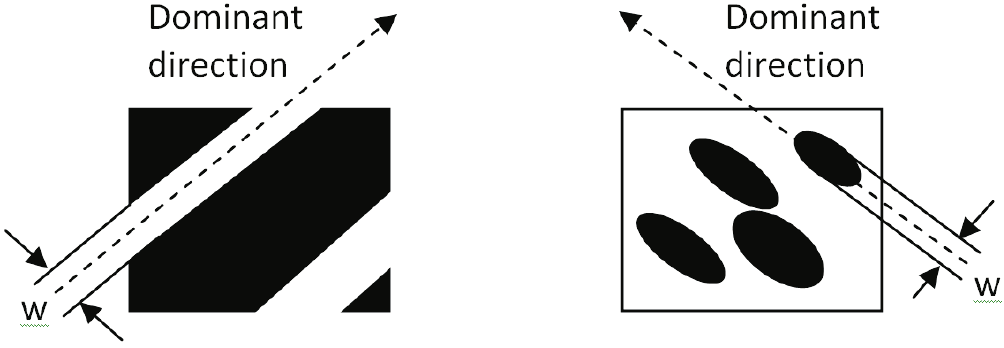}
\end{tabular}
\caption{The minimal width of foreground features in the direction perpendicular to dominant orientation can serve as a granularity index in formulation 2 from Fig.~\ref{Fig:ElevationFormulations}.
\label{Fig:ElevationForm2}}
\end{center}
\end{figure}

\begin{figure}[t]
\begin{center}
\includegraphics[width=0.55\textwidth]{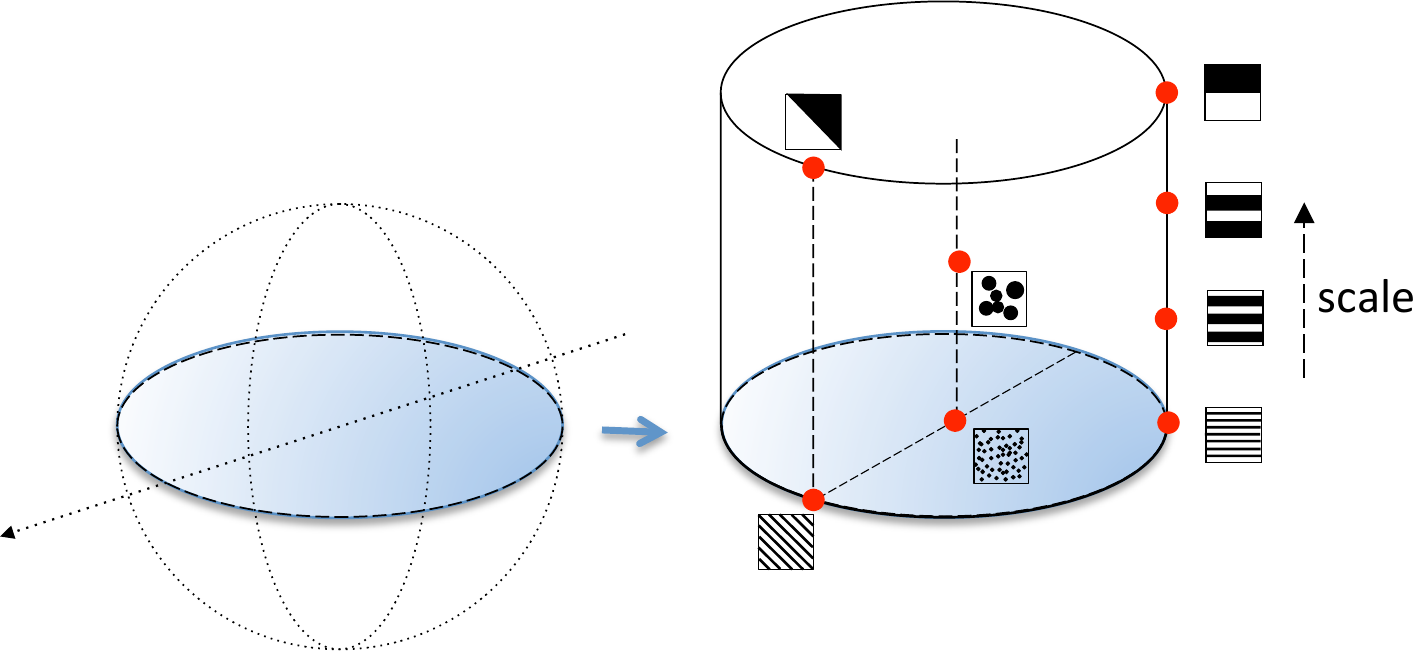}
\caption{Scale cylinder adds a fourth dimension along which the scale of patterns changes keeping other parameters fixed.\label{Fig:ScaleCylinder}}
\end{center}
\end{figure}

\begin{figure}[t]
\begin{center}
\includegraphics[width=0.4\textwidth]{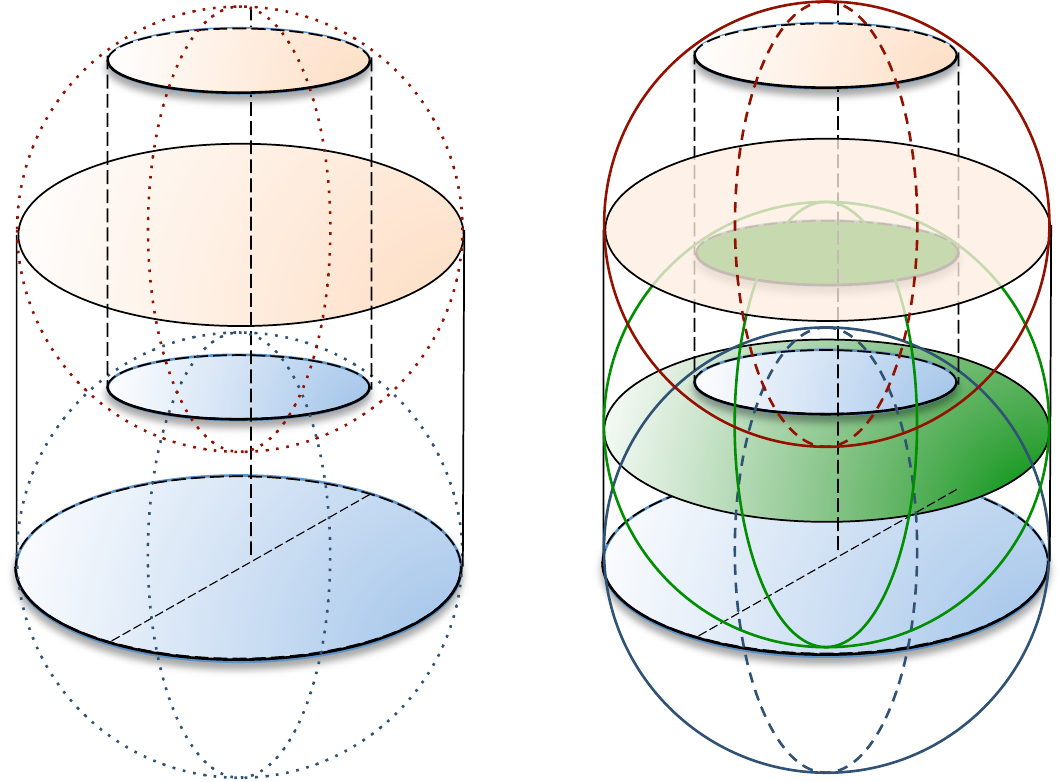}
\caption{Construction of a scale hyper-ball from scale cylinders.\label{Fig:ScaleHypesphere}}
\end{center}
\end{figure}

\begin{figure}[t]
\begin{center}
\includegraphics[width=0.55\textwidth]{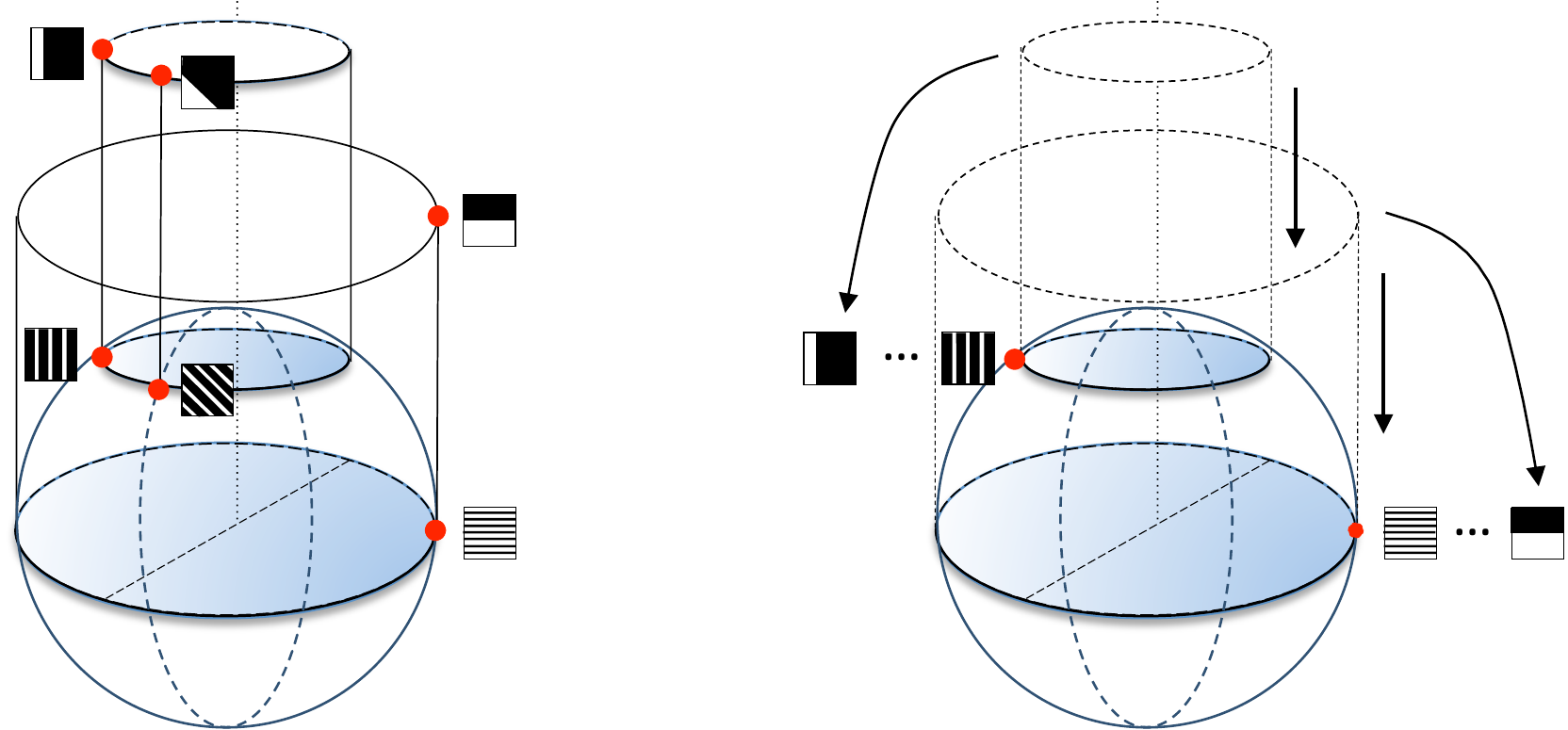}
\caption{Pooling out scale cylinders unfolds the scales in an extra dimension. Collapsing the cylinders back to their bases projects 4D encoding back to 3D where patterns at different scales map to the same point.\label{Fig:ScaleExtendCollapse}}
\end{center}
\end{figure}

Two resulting pattern encoding schemes
with different formulations of the balance between dark and bright tones
are depicted in Fig.\ref{Fig:ElevationFormulations}. In the first formulation (Fig.\ref{Fig:ElevationFormulations}, left) the elevation angle $\chi$ is associated with the \emph{mean patch intensity}, and in the second formulation (Fig.\ref{Fig:ElevationFormulations}, right), the elevation is associated with a \emph{granularity index}, which can be understood as the size of
the smallest foreground region in the direction perpendicular to the
dominant direction (see
Fig.~\ref{Fig:ElevationForm2}).

Notice that different image patterns map to the same point as long as they have the same level of regularity, the same dominant direction and the same mean grey value or the same granularity index. In the mean intensity formulation of the elevation angle, patterns with many thin lines and those with few thick lines of the same orientation can map to the same point. All entirely random and anisotropic noise patterns map
to the central axis (or to the same point on this axis for a fixed mean intensity), irrespective of their scale or the size of blobs. These can be separated in an extra dimension that encodes the scale. In the second formulation, patterns with different mean intensities can map to the same point if their granularity index is the same.

In our practical encoding tool, we will associate the elevation angle to the mean pattern intensity (as in formulation 1 from Fig.~\ref{Fig:ElevationFormulations}). In the following, one possible construction is
described that extends this 3D pattern encoding scheme to
four dimensions, where the notion of scale is made explicit.

Suppose we take an arbitrary intersection of the Poincar\'{e}
sphere with a plane parallel to the equatorial plane. The resulting circle with its interior is a locus of points that represent all patterns
with the same
mean intensity and with all possible dominant orientations and
regularity levels. Without loss of generality, in Fig.~\ref{Fig:ScaleCylinder} we take the disk
that is exactly on the equatorial plane. Each point on this disk
is a projection of a line from a 4-dimensional space, which encodes patterns at different
scales, all sharing the same mean intensity, the
same dominant orientation and the same level of regularity.
Equivalently, this means that we can unfold the points
corresponding to different scales, by extending each disk in a
plane parallel to the equatorial plane to a cylinder.

Suppose now that we have pulled out the corresponding cylinders from all the disks in planes parallel to the equatorial plane and we take from each of these cylinders a cross section at the same distance $s$ from the cylinder base. The resulting disks make a new ball, corresponding to the scale $s$. A collection of all the balls for all scales constitutes a hyper-ball, as illustrated in Fig.~\ref{Fig:ScaleHypesphere}. 

It is useful to imagine that the scale cylinders can be at any moment and at any plane parallel to the equatorial plane pulled out and collapsed back to their base. In this way we can switch between the 4D space where scales have explicit meaning and its projection onto a 3D space where each point corresponds to a variety of patterns at different scales (see an illustration in Fig.~\ref{Fig:ScaleExtendCollapse}). In the remainder of the paper, we construct a practical tool for the 3D case, having on mind that the scales can be unfolded and separated, if desired, as it was discussed here.

\section{A practical pattern encoding tool} \label{Sec:ToyEncodingScheme}

\begin{figure}[b]
\begin{center}
\includegraphics[width=0.4\textwidth]{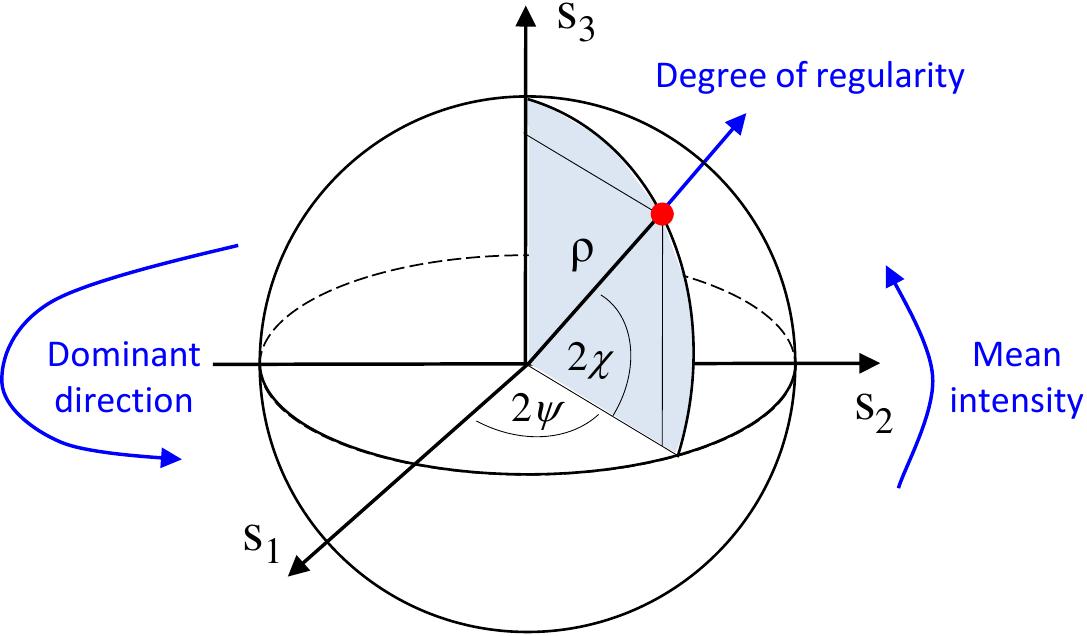}
\caption{Features used in our toy encoding tool. \label{Fig:ToyEncodingParameters}}
\end{center}
\end{figure}

Here we construct a concrete graphical tool for encoding visual patterns based on the ideas described above. 
Because of its simplicity in terms of feature extraction, it can be considered as a toy encoding tool.
Fig.~\ref{Fig:ToyEncodingParameters} shows the meaning of the three chosen parameters: the degree of regularity, the dominant orientation and the mean intensity. In the following we design simple methods to estimate each of these three parameters.

\subsection{Encoding azimuthal angle by dominant direction}\label{Sec:DominantOrientation}

Estimating dominant direction (orientation) of visual patterns is a widely studied subject \cite{Jafari-Khouzani2005, Jiang05, Kass87} and we could  plug in here any of the existing methods. A reader who is less interested in our particular orientation estimation technique can skip this subsection and simply assume another orientation estimation method.

\begin{figure}[t]
\begin{center}
\includegraphics[width =0.4\textwidth]{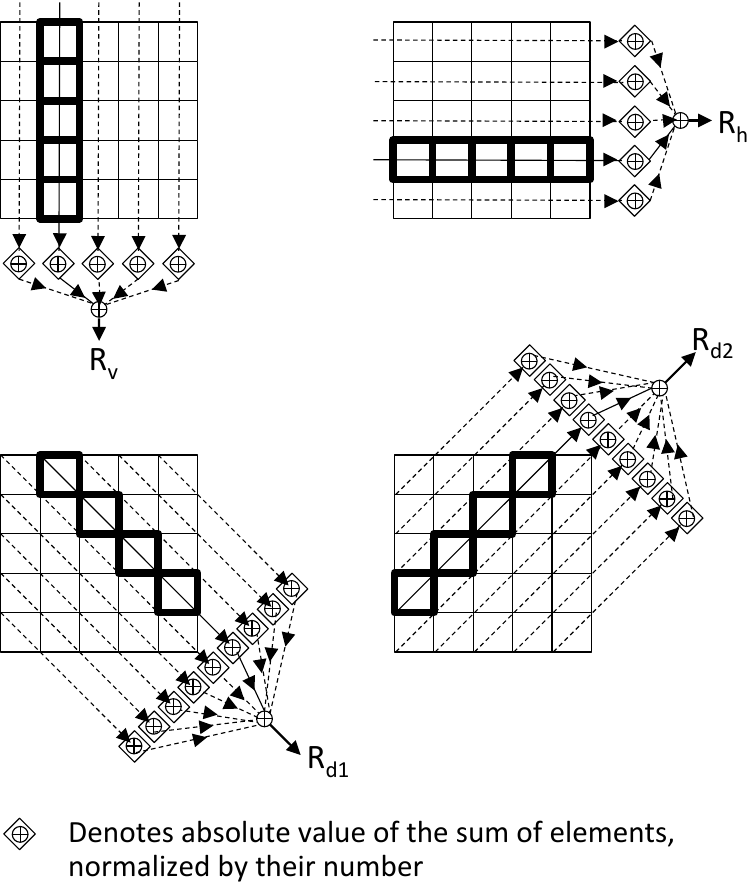}
\caption{Four 
projectors $R_k$, $k \in \{v,h,d1,d2\}$ (subsequently normalized to the $r_k$'s from Fig.~\ref{Fig:DominantDirEst}) in the proposed orientation estimation method. \label{Fig:FourProjectors}}
\end{center}
\end{figure}

We define here a method of estimating the dominant orientation of an image patch, based on four Radon-like projectors, illustrated in Fig.~\ref{Fig:FourProjectors}. Each of these projectors, when applied to a \emph{zero mean} image patch,  gives a single number that characterizes the degree of directionality in a particular direction.

 Let $\mathbf{w}=\{w_{i,j}\}$, $i=1,...,M$,
$j=1,...,N$ denote a zero mean patch derived from some image patch
$\mathbf{I}=\{I_{i,j}\}$ with mean intensity $\mu_I$, as 
$w_{i,j}=I_{i,j}-\mu_I$. In each of the projectors from
Fig.~\ref{Fig:FourProjectors}, the pixel values are first summed
along a given projection ray  and
divided by the number of elements along that ray. The absolute
values of these ray-wise normalized Radon projections are then
averaged along all the rays. For example, the horizontal projector
$R_h$ averages the absolute values of the row-wise projections
$\sum_{j=1}^Nw_{i,j}$ over all the rows $i$. Equivalently, the
vertical projector $R_v$ averages the absolute values of
column-wise projections $\sum_{i=1}^Mw_{i,j}$. The rationale
behind this is very simple: Suppose we have dominantly vertical
features (stripes); the row-wise projections $\sum_{j=1}^Nw_{i,j}$
will be close to zero because $w_{i,j}$ will have alternating
signs along each row; the column-wise projections
$\sum_{i=1}^Mw_{i,j}$ will be large positive and large negative
numbers (because $w_{i,j}$ will be mostly of the same sign along
each column $j$). By summing the absolute values of these row-wise
and column-wise projections, we promote the projection direction
that aligns better with the dominant direction in the image patch
$\mathbf{w}$.

Formally, let $U_k(l)$ denote a \emph{normalized} discrete Radon
transform of a zero-mean image patch $\mathbf{w}$:
\begin{equation}\label{Eq:nDRT}
U_k(l)=\frac{1}{|L_{k,l}|}\sum_{(i,j) \in L_{k,l}}w_{i,j}
\end{equation}
where $L_{k,l}$ denotes the set of points that make up a digital
line segment (finite ray) with the slope index $k$ and the
intercept index (ray number) $l$. This formulation is, apart from the normalization term $|L_{k,l}|$ equivalent to the classical discrete Radon transform\footnote{This should not be confused with the \emph{finite} Radon transform, commonly defined after
\cite{Matus93} for square images $p \times p$, where $p$ is
prime and the image is assumed to be periodic with $p$ in both
horizontal and vertical directions by defining $p \times p$ array
as the finite group $Z^2_p$ under addition. In that case, there
are $p+1$ distinct line directions: $L_{k,l}=\{(i,j) : j=ki+l
(\textrm{mod} p ),  i \in Z_p \}$,  $0 \leq k \leq p$ and
$L_{p,l}=\{(l,j) : j \in Z_p \}$, where all the finite lines are
of equal length $p$ because they have a wrap around effect (see
also \cite{Do03, Kingston07}). In our case the wrap around effect is not
desirable and we rather define $L_{k,l}$ as strips of variable length along the
patch, as in \cite{Gotz96, Brady98, Press06}.} \cite{Gotz96, Brady98, Press06}, but in our case the normalization factor is very important.

Define now the $k$-th projector as
\begin{equation}
R_{k}=\frac{1}{\rho_k}\sum_{k=1}^{\rho_k}|U_k(l)|
\end{equation}\label{Eq:Projector}
with $\rho_k$ being the number of rays with slope index $k$. For practical computation of $R_k$ one could employ any algorithm for  discrete Radon transform to find $U_k(l)$, like the fast algorithms from \cite{Gotz96, Brady98, Press06}, slightly updated to take into account the normalization term $|L_{k,l}|$. Since we here need only projections in four special directions $k \in \{h, v, d1, d2\}$ (see Fig.~\ref{Fig:FourProjectors}), we do not need to use any sophisticated algorithm for the discrete Radon transform, but we can easily derive the expressions. Obviously $R_h=\frac{1}{MN}\sum_{i=1}^{M}\Bigl|\sum_{j=1}^{N}w_{i,j}\Bigr|$ and $R_v=\frac{1}{MN}\sum_{j=1}^{N}\Bigl|\sum_{i=1}^{M}w_{i,j}\Bigr|$. The expressions for $R_{d1}$ and $R_{d2}$ along with the derivation are given in the Appendix. Let us normalize the resulting projectors such that their squares in horizontal and vertical direction sum to 1:
\begin{equation}\label{Eq:r_hv}
r_{h,v}=\frac{R_{h,v}}{\sqrt{R_h^2+R_v^2}}
\end{equation}
and analogously for the two diagonal directions:
\begin{equation}\label{Eq:r_d1d2}
r_{d1,d2}=\frac{R_{d1,d2}}{\sqrt{R_{d1}^2+R_{d2}^2}}
\end{equation}

\begin{figure}[t]
\begin{center}
\includegraphics[width =0.3\textwidth]{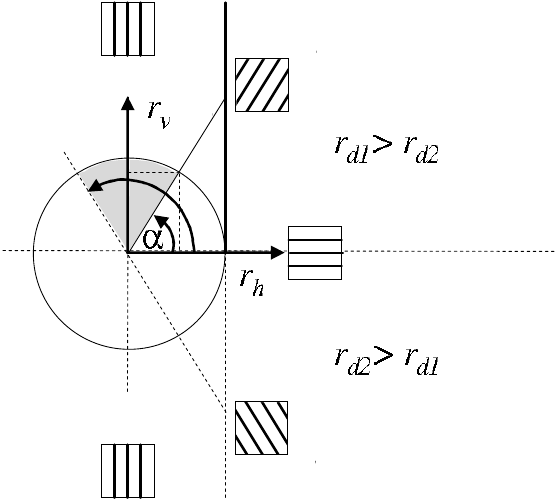}
\caption{Estimating pattern orientation from the four normalized Radon-like projectors. \label{Fig:DominantDirEst}}
\end{center}
\end{figure}

Note that for ideally horizontal patterns, $r_h=1$ and $r_v=0$, while the opposite is true for the vertical patterns. Hence, we can estimate the dominant direction of an image pattern according to the scheme from Fig.~\ref{Fig:DominantDirEst}. Since the projectors $r_h$ and $r_v$ defined in
(\ref{Eq:nDRT})-(\ref{Eq:r_hv}) are always positive,
$\arctan(r_v/r_h)$ is within $[0,\pi/2]$. The value of
$\alpha=\arctan(r_v/r_h)$, $\alpha\in[0,\pi/2]$ determines the
dominant direction up to mirroring ambiguity (see
Fig.~\ref{Fig:DominantDirEst}). This ambiguity can be resolved by
investigating the value of $r_{d1}$ relative to that of $r_{d2}$
and applying if necessary the adequate correction:
\begin{equation}\label{Eq:angle}
\psi=\Bigl\{
\begin{array}{ll} \arctan{(r_v/r_h)}, & \
\mathrm{if} \ \ r_{d1} \geq r_{d2}, \\  \pi - \arctan{(r_v/r_h)},  & \
\mathrm{otherwise.}
\end{array}
\end{equation}

\begin{figure*}[t]
\begin{center}
\includegraphics[width=\textwidth]{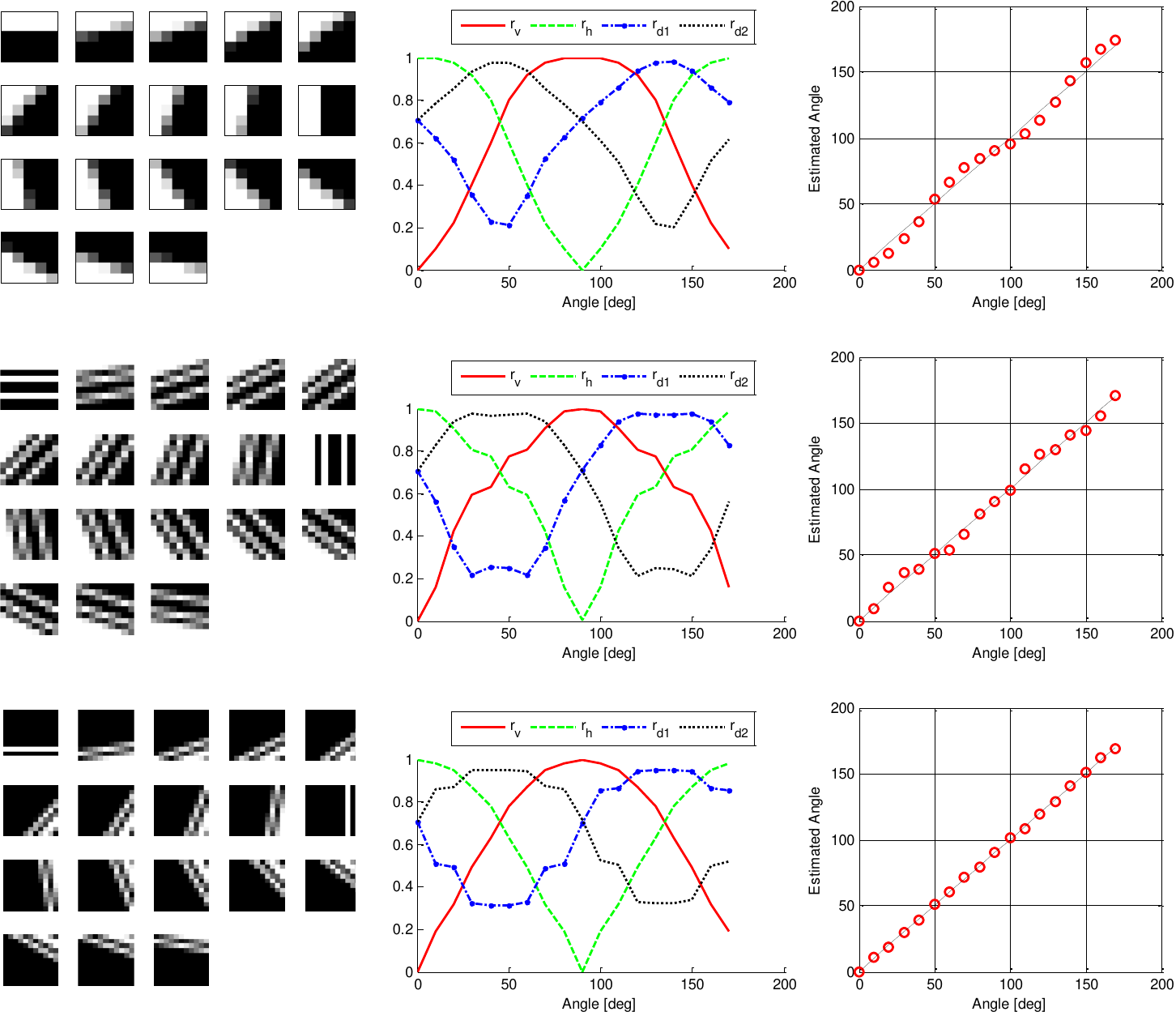}
\caption{Examples of dominant orientation estimation using the method of four
normalized Radon-like projectors. {\bf Left:} a sequence of rotated image patches with ideally parallel lines, where the orientation angle ranges from 0 to 170$^\circ$ in steps of 10$^\circ$. The rotation was implemented using bilinear interpolation. {\bf Middle:} the evolution of the four projectors $r_h$, $r_v$, $r_{d1}$ and $r_{d2}$ with the orientation angle. {\bf Right:} the estimated angle versus the actual angle of the dominant pattern orientation. From top to bottom, the patch size is 5$\times$5, 9$\times$9 and 11$\times$11.  \label{Fig:DirEstimExample1}}
\end{center}
\end{figure*}

\begin{figure*}[t]
\begin{center}
\includegraphics[width=\textwidth]{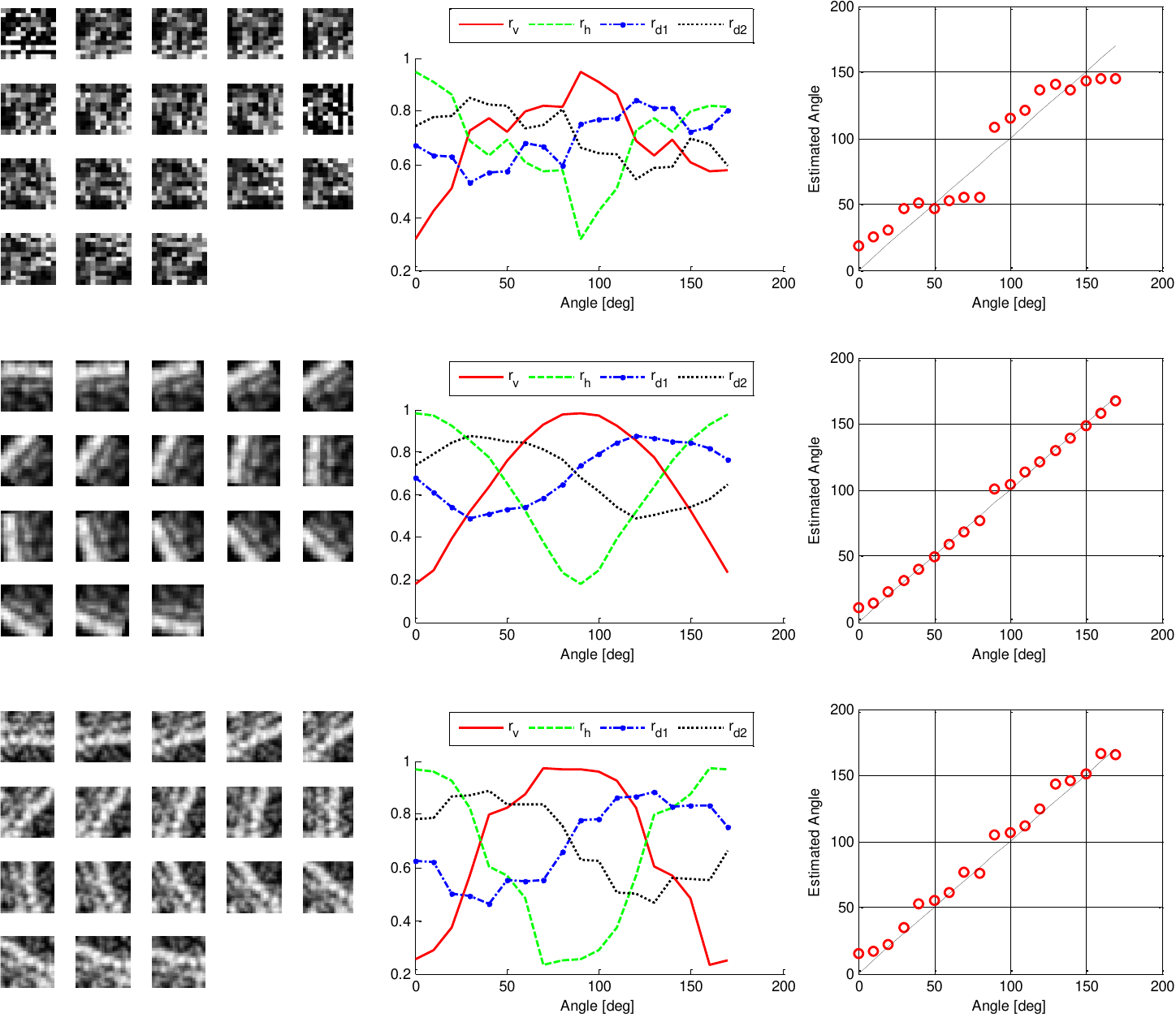}
\caption{Same as in Fig.~\ref{Fig:DirEstimExample1}, but for degraded image patches affected by noise and blur. From top to bottom, the patch size is 11$\times$11, 15$\times$15 and 21$\times$21, and in all cases SNR=0dB. \label{Fig:DirEstimExample2}}
\end{center}
\end{figure*}

A few remarks: note that we do not actually need the exact values of $r_{d1}$ and $r_{d2}$, so it might be sufficient to compute only $r_{h}$ and $r_{v}$ using (\ref{Eq:nDRT})-({\ref{Eq:r_hv}) and to devise a fast method to check whether $r_{d1} \geq r_{d2}$, which would make the whole procedure even simpler. Given that even now the computation of $r_{d1}$ and $r_{d2}$ with the expressions given in the Appendix is quite simple to implement and fast, we will not bother here to simplify it further. It should also be noted that (\ref{Eq:angle}) gives the same estimates $\psi$ if the normalization in (\ref{Eq:r_hv}) and (\ref{Eq:r_d1d2}) is replaced by $r_k=R_k/\sqrt{\sum_j R_j^2}$, $j,k \in \{h,v,d1,d2\}$. The normalization in (\ref{Eq:r_hv}), (\ref{Eq:r_d1d2}) is easier to link visually to the scheme in Fig.~\ref{Fig:DominantDirEst}, where the circle is then trigonometric circle with $r_h^2+r_v^2=1$.

Fig.~\ref{Fig:DirEstimExample1} illustrates the performance of this method on sequences of rotated image patches of different sizes. These test patches were generated starting from a binary image patch with ideal horizontal lines and then rotated in steps of 10 degrees using bilinear interpolation. In the case of these degradation-free image patches the evolution of the four normalized projectors $r_h$, $r_v$, $r_{d1}$ and $r_{d2}$  with the rotation angle (middle of Fig.~\ref{Fig:DirEstimExample1}) shows a perfect symmetry (the values are exactly the same for angles $\psi$[deg] and 180$^{\circ}$-$\psi$[deg]). 
Note that the patch size in the top row is only $5 \times 5$, so this simple method is applicable to even very small image patches. Fig.~\ref{Fig:DirEstimExample2} demonstrates the performance in the presence of noise and blur. In this case, we also started from a degradation-free image patch with ideal horizontal lines, which was then degraded and rotated in steps of 10 degrees. For example, patches in the top of Fig.~\ref{Fig:DirEstimExample2} are degraded versions of those from the bottom of Fig.~\ref{Fig:DirEstimExample1} (with signal-to-noise-ratio $SNR=0dB$). Now the evolution of the four projectors with respect to the orientation angle no longer shows the symmetry that was present in patches with ideally parallel lines, while by normalization, it is still ensured that $r_h^2+r_v^2=1$ and $r_{d1}^2+r_{d2}^2=1$. The method is rather robust to noise and blur (see the situation in the top of Fig.~\ref{Fig:DirEstimExample2} with extremely noisy patches) and this can be attributed to the embedded averaging operations. 

Comparing this estimator with available ones is outside the scope of this paper, since our focus here is not on the particular methods for extracting the features; any of the existing methods for dominant pattern orientation could be plugged in as well in our pattern encoding scheme. Studying more general use of the here presented method for dominant orientation estimation in comparison with state-of-the-art methods could be interesting, but here this would make a too big depart from the main scope of the paper so we leave it for future investigation.

%


\subsection{Encoding radial distance by pattern regularity}

Estimating regularity of visual patterns and textures is another fundamental problem in computer vision. We can use for our encoding tool any of the existing regularity estimation methods as long as they express pattern regularity as a single number between 0 and 1. Here we experiment with two simple methods, based on patch entropy and on consistency of local dominant directions.

\subsubsection{Entropy based pattern regularity measures}

Let $\mathbf{h}=[h_1,...,h_n]$ denote the normalized histogram of a grey scale image patch, such that $\sum_{i=1}^n h_i=1$ and $h_i>0$ (i.e., the empty bins are discarded or the bin size is adjusted such that all bins are populated). The patch entropy
\begin{equation}\label{Eq:Entropy}
E=-\sum_{i=1}^n h_i \log_2(h_i)
\end{equation}
increases as $\mathbf{h}$ becomes more uniform, being zero for a uniform patch and reaching maximum when all the possible pixel values are equally often present (e.g., $E=8$ for a greyscale patch with intensities in \{0,...,255\} and where each grey value appears equally often). The entropy definition as such is ignorant to spatial structure: reshuffling arbitrarily pixel intensities in an image patch has no effect on its entropy. For this reason, entropy-based methods for estimating pattern regularity should first apply a kind of data partitioning which takes care of the spatial structure, as was suggested in \cite{Okada08}. In particular, the method of \cite{Okada08} proceeds in three steps: 1) stratifying, i.e.,
partitioning data domain, 2) deriving an intensity distribution
for each stratum and 3) computing set-similarity of the
distributions using Jensen-Shannon divergence \cite{Lin91}.

If we are interested in relatively small image patches, a simple regularity measure based on patch entropy alone may be sufficient (a small, noise-free image patch, which consists e.g. of parallel stripes will typically contain only few distinct grey values in practice and thus will have a low entropy, whereas a patch affected by noise and/or blur will have a wider spread of grey values resulting in a larger entropy). A possible mapping from patch entropy (\ref{Eq:Entropy}) to a regularity level $\rho \in [0,1]$ for grey scale patches is: 
\begin{equation}\label{Eq:RHO_entropy}
\rho_E=\min(1-\frac{E-1}{7},1)
\end{equation}
where subscript $E$ denotes that this regularity estimate is entropy-based.
In practice,  we keep in the histogram $\mathbf{h}=[h_1,...,h_n]$ only the bins that have counts above some threshold (10\% of the maximum count). We find that this regularity estimate works reasonably well on most of the relatively small image patches (with sizes $15 \times 15$ or smaller) and even on larger ones when the underlying patterns are relatively simple (e.g., line like). The use of this regularity measure in our pattern encoding tool is illustrated later (in Section \ref{Sec:PatternEncodingExamples}). For general patterns, measures like \cite{Okada08} should be considered instead.

\subsubsection{Local directional consistency (LDC)}

If we define regular patches as fully oriented ones, we can think of regularity as a kind of local directional consistency (LDC), which describes how well local dominant directions agree. A pattern that consists of parallel stripes will have
consistently the same dominant orientation in each window of
sufficient size. So, we can divide the patch into $S$ sub-patches
(using a sliding or not sliding window depending on the patch
size\footnote{Using a sliding window will result in more counts in
the histogram and hence in a more accurate estimation. If the
patch size is large relative to the window size, then we will
have enough measurements (counts in the histogram) even without
sliding, so in that case, for computational complexity reasons, it
is more practical to use non-overlapping windows.}) and
calculate the dominant orientation angle $\psi_i$ in each window $i$.
The spread of the histogram $\mathbf{h}_{\psi}$ of the block-wise angles
$\{\psi_1,...,\psi_S\}$ gives us an idea about the uniformity of
the present edge directions within the patch and hence about its
regularity. The smaller the spread of the histogram $\mathbf{h}_{\psi}$,
the larger regularity $\rho$.

Since $\rho\in[0,1]$, we will demand that $\rho=1$ if the
histogram $\mathbf{h}_\psi$ contains only one peak (i.e., only one
populated bin) and $\rho=0$ if the histogram spreads over the
whole range $[0,180^\circ]$. Denote by $B$ the total number of bins in the histogram $\mathbf{h}_\psi$
and by $b$ the number of populated bins that contain counts above some small
threshold $T_{hr}$ (typically, we set $T_{hr}$ to 5\% of the
maximum count in a bin). We define a LDC-based patch regularity measure as:
\begin{equation}
\rho_{LDC}=\frac{B-b}{B-1}
\end{equation}
In case where $b=B$ (all bins populated) $\rho_{LDC}=0$ and in case
where $b=1$ we have $\rho_{LDC}=1$, conforming to the requirements
above. One could think of a more
sophisticated approach that takes into account the shape or entropy of the local orientation
histogram. Fig.~\ref{Fig:BlockwiseHists} illustrates examples of block-wise local orientation histograms $\mathbf{h}_\psi$ for some artificial test patterns and real image patches.

This regularity measure is applicable only if the regularity is treated as the degree of pattern orientedness (which is the case in our toy encoding tool, but may not be the case in general). The main limitation of this method is that it relies on computing local directions in sub-blocks, which makes it impractical if the patch under investigation is rather small (in practice reasonable results can be obtained only for patches that are at least 7$\times 7$ pixels).

\begin{figure}[t]
\begin{center}
\begin{tabular}{cc}
\includegraphics[width=0.45\textwidth] {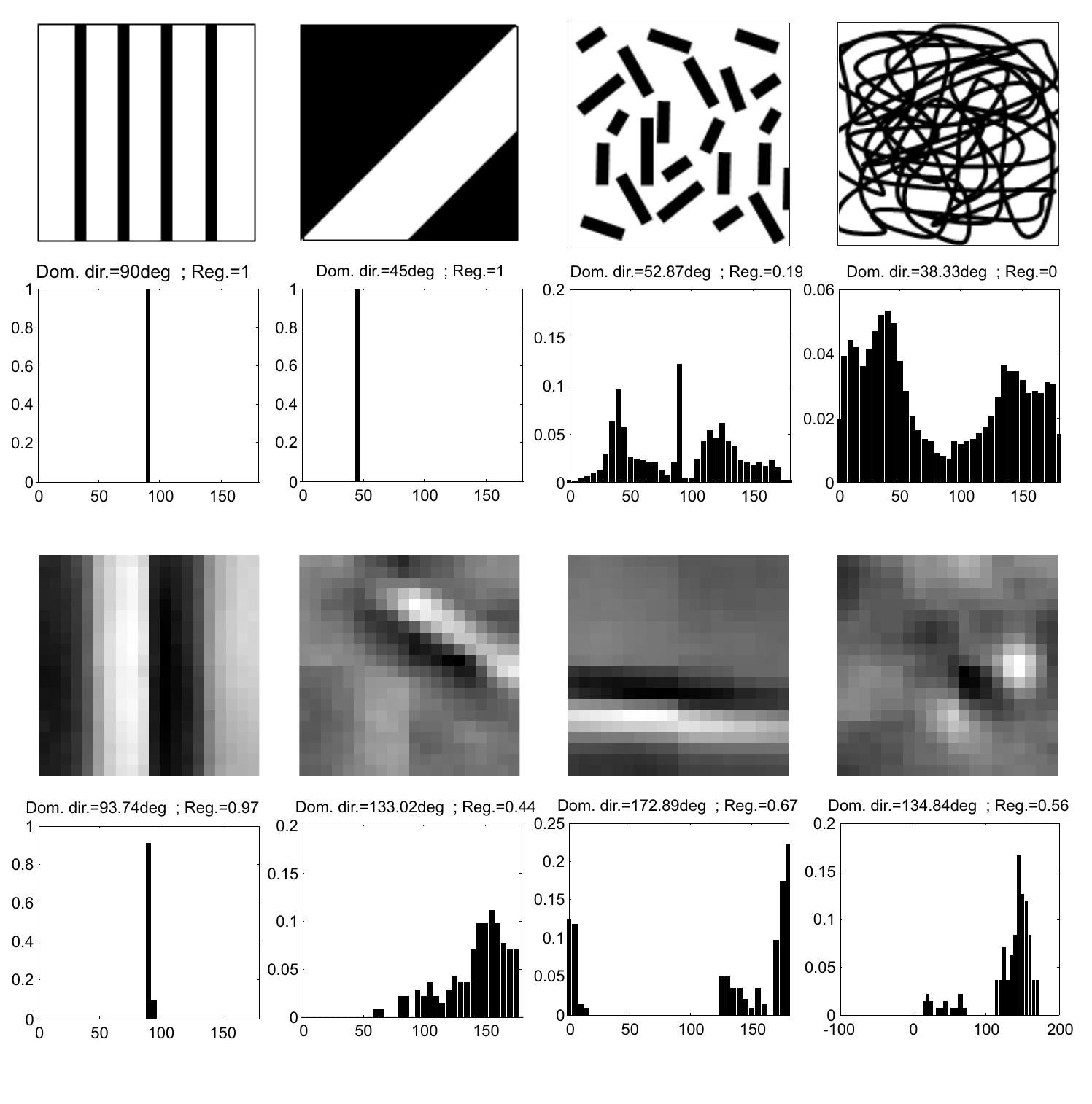}
\end{tabular}
\caption{Block-wise direction histograms $\mathbf{h}_\psi$
 for some artificial and real image patches (with block size $8 \times 8$). \label{Fig:BlockwiseHists}}
\end{center}
\end{figure}

\subsection{Encoding elevation angle by mean pattern intensity}

Here we assign the elevation angle to mean pattern intensity. Alternatives could include a granularity index as in the conceptual scheme in Fig.~\ref{Fig:ElevationFormulations}, right.

Let $\theta=2\chi$,
$\theta\in[-\pi/2,\pi/2]$ encode the ``whiteness'' or the
``blackness'' of the patch, so that $\theta=-\frac{\pi}{2}$
corresponds to black and $\theta=\frac{\pi}{2}$ to white. 
For an $M \times N$ image patch $\mathbf{I}=\{I_{i,j}\}$ with $L$ intensity levels, $I_{i,j} \in \{0,L-1\}$, the normalized mean intensity 
\begin{equation}
T=\frac{\sum_{i=1}^{M}\sum_{j=1}^{N}I_{i,j}}{LMN}
\end{equation}
is in the range $[0,1]$ and a linear mapping
\begin{equation}\label{Eq:NormalizedIntensity}
\theta=2\chi=(T-0.5)\pi
\end{equation}
translates $T$ values into desired elevation angles $\theta=2\chi\in[-\pi/2,\pi/2]$.
This completes one possible specification for pattern
encoding via the three spherical coordinates $\rho$ (regularity),
$2\psi$ (dominant angle) and $2\chi$ (mean intensity), which agrees with the conceptual scheme from
Fig.~\ref{Fig:ElevationFormulations} (left). We show next examples of encoding image patches and learned image atoms using this tool.

\begin{figure}[t]
\begin{center}
\begin{tabular}{cc}
\includegraphics[width =0.6\textwidth]{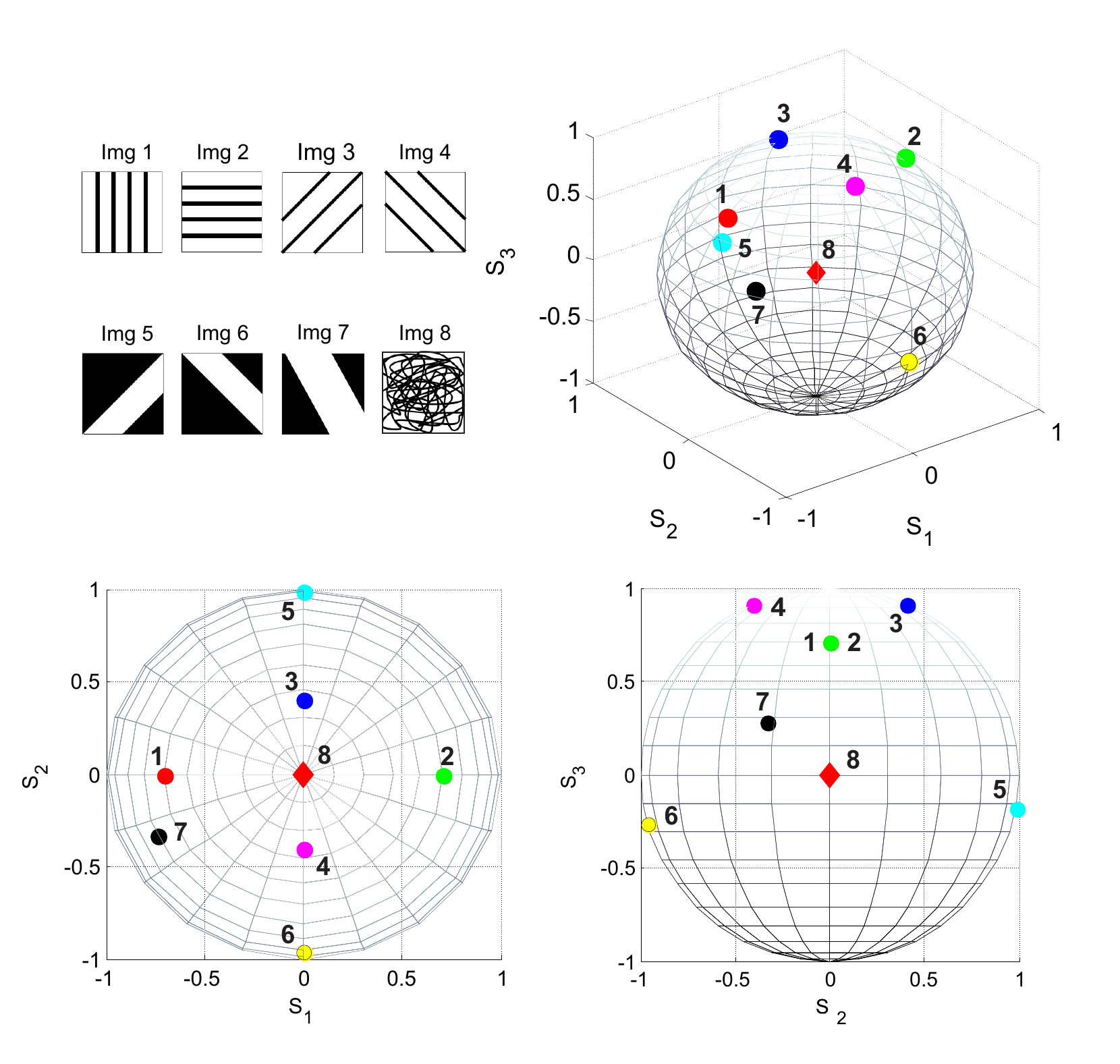}
\end{tabular}
\caption{Pattern encoding illustrated on eight
artificial image patches. \label{Fig:CodeExample}}
\end{center}
\end{figure}

\begin{figure}[t]
\begin{center}
\begin{tabular}{cc}
\includegraphics[width=0.75\textwidth]{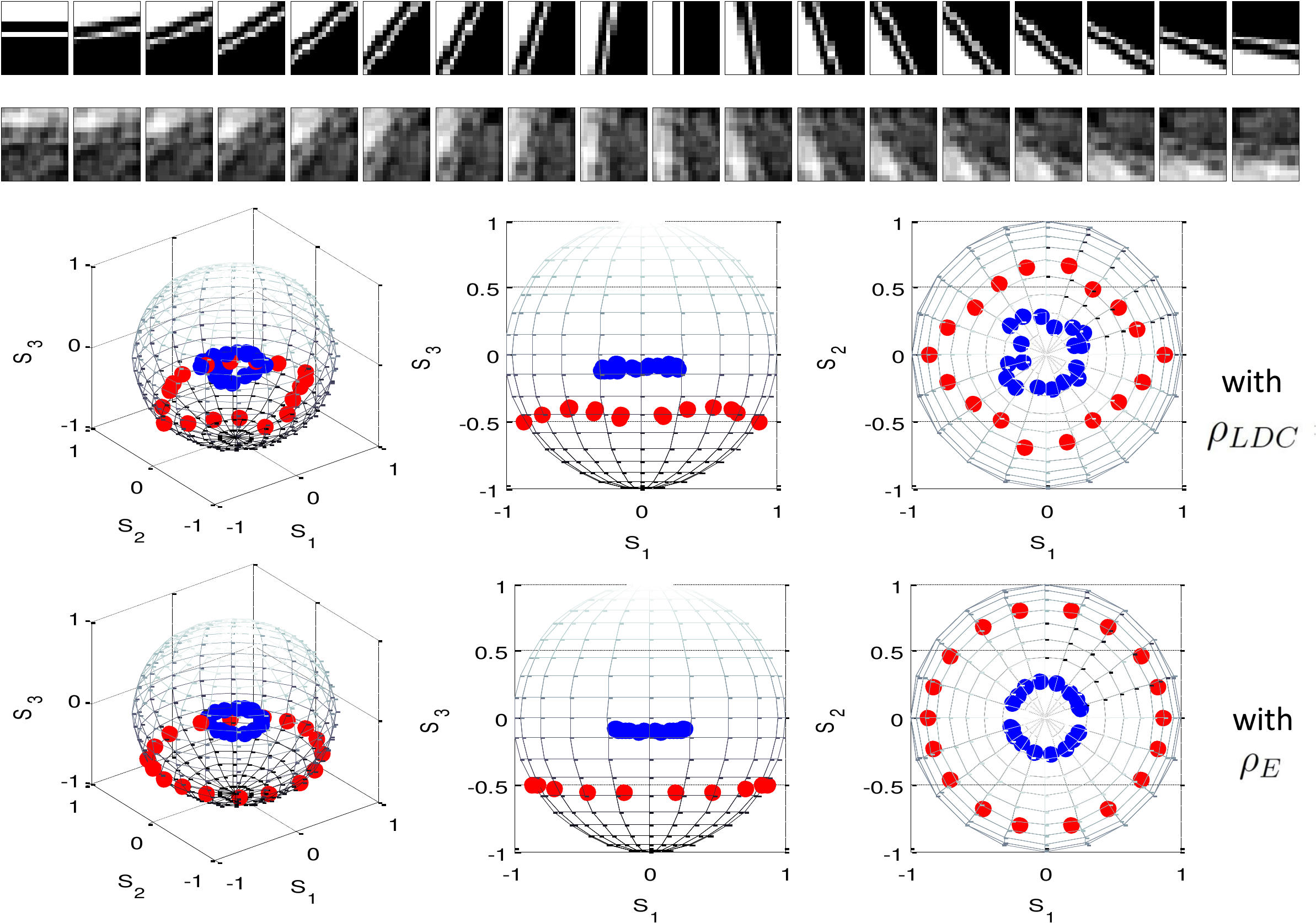}
\end{tabular}
\caption{Pattern encoding illustrated for two sequences of image patches: ideal directional (top row) - red dots, and degraded (second row) - blue dots. Two variants of pattern regularity estimation are employed: LDC-based, (third row), and entropy-based (fourth row).
\label{Fig:PatternEncodingExample}}
\end{center}
\end{figure}

\subsection{Pattern Encoding Examples} \label{Sec:PatternEncodingExamples}

Even though we can store the points in the spherical coordinates, it is usually more practical to compute first the three corresponding coordinates in the Cartesian system, that will actually correspond to the three Stokes parameters of a visual pattern\footnote{In optics, $\rho$ in these equations would be denoted by $S_0$, which is the fourth Stokes parameter.} (see Fig.~\ref{Fig:PolarizationEllipseSphere}):
\begin{eqnarray}\label{Eq:StokesPoincare}\label{Eq:SphericalToCartesian}
S_1=\rho\cos(2\chi)\cos(2\psi) \nonumber
\\
S_2=\rho\cos(2\chi)\sin(2\psi) \\
S_3=\rho\sin(2\chi) \nonumber
\end{eqnarray}

Fig.~\ref{Fig:CodeExample} illustrates the proposed encoding
scheme on a set of eight artificial patches. Seven of these are
fully regular (ideally directional or fully polarized) and are assigned to seven corresponding
points on a sphere with radius $\rho=1$. The eighth patch has no
clear dominant direction and is positioned in the centre of the
sphere. The regularity was estimated as the local directional consistency $\rho_{LDC}$.

Another example in Fig.~\ref{Fig:PatternEncodingExample} shows encoding of two sequences of image patches: a sequence of degradation-free fully oriented patches (top row) and a sequence of noisy and blurred patches (second row). In both of these two sequences all the patches are obtained by rotating the same patch from 0$^\circ$ to 170$^\circ$ in steps of 10$^\circ$. The results are shown both for the LDC-based and for the entropy based regularity estimation. With both encoding schemes the two distinct sequences of image patches map to two distinct point constellations. When $\rho_E$ was used, these point constellations are on two almost perfect circles. The circle corresponding to the ideally oriented degradation-free patches (red dots) is on the surface of the unit sphere and the other one (blue dots), corresponding to the degraded patches is of a smaller radius. In the case where $\rho_{LDC}$ is used, the two point constellations although not on ideal circles are still two distinct ring-like structures. The ring of points formed by the degraded patches is near to the equatorial plane because the normalized mean intensity $T$ of these patches is nearly 0.5, while the ring corresponding to the non-degraded patches is in the lower hemisphere, because the mean intensity of these patches is lower (black tone dominates over the white).

\section{Example Applications} \label{Sec:Applications}

In this Section, three possible applications are presented to demonstrate the potentials of the developed pattern encoding tool. Further work will be needed to devise practical methods and to validate them in concrete image processing and computer vision applications, which falls out of the scope of this paper. 


\begin{figure*}[t]
\begin{center}
\begin{tabular}{cc}
\includegraphics[width=0.75\textwidth]{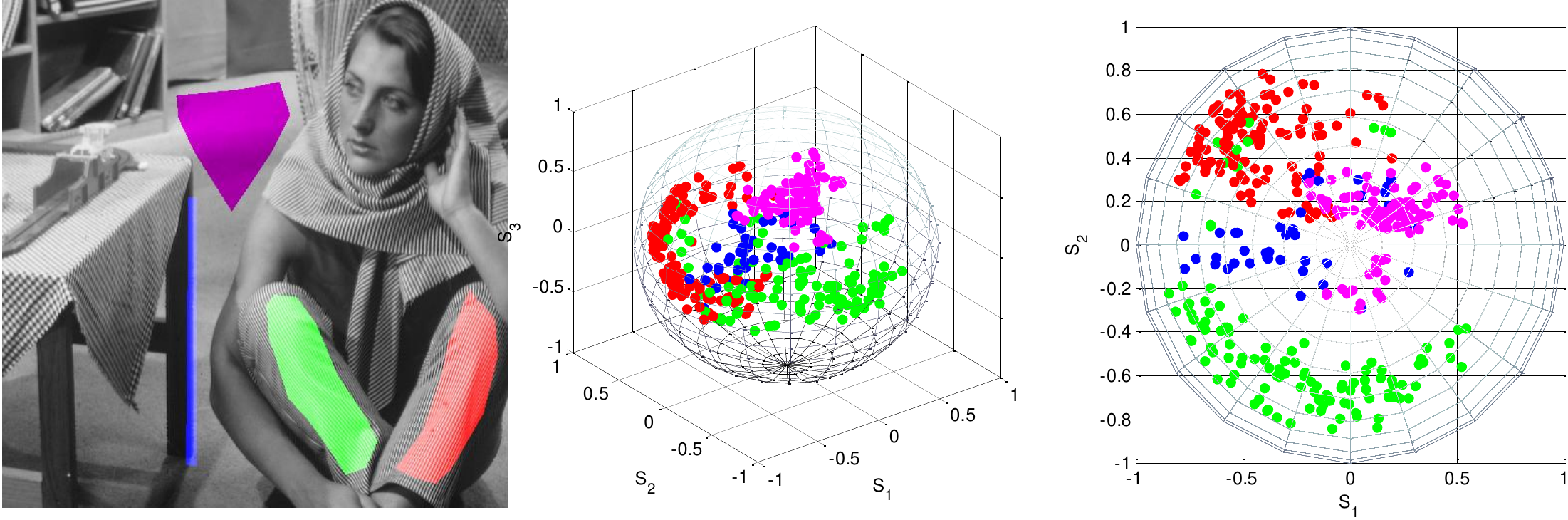}
\end{tabular}
\caption{An example of patch clustering. Four regions were manually selected and randomly chosen patches from these regions were encoded as dots of the matching color.
\label{Fig:PatchClustering}}
\end{center}
\end{figure*}

\subsection{Patch Clustering}

Obviously, the proposed graphical tool can be used to visualize clusters of image patches that share some common properties, like dominant orientation, the degree of regularity and mean intensity. Fig.~\ref{Fig:PatchClustering} shows one example, where a user has selected four image regions with different textural characteristics. From each of these regions a number of patches were randomly selected and visualized as dots of the matching color. The resulting point clouds are located in regions with the corresponding dominant orientations. The point cloud that corresponds to nearly uniform patches in the background is located around the central axis which indicates lack of any pronounced dominant orientation. The patches taken from a narrow region in the area of the table leg are correctly encoded as points in the region corresponding to dominant vertical orientation. Similar holds for two regions on the trousers. There, the spread of the local dominant orientations is larger, hence the corresponding point clouds are more smeared but they are still clearly centred around the globally dominant orientations in these regions (approximately 60$^\circ$ and 120$^\circ$).

This type of pattern encoding can be useful as a pre-processing step in searching for self-similar image regions, e.g. for non-local image denoising or inpainting methods mentioned in the Introduction. It could be also interesting to explore its use in content description: for example, randomly taken patches from images of urban scenes with dominantly horizontal and vertical orientations will have a different representation than those taken from images of natural textures (sand, water, sky, etc) which have no clear dominant orientation and which appear less regular).

\subsection{Dictionary encoding example}
\label{Sec:ExperimentalResults}

Another possible application of the developed pattern encoding tool is in visualizing the properties of dictionaries of learned image atoms, and assisting in this way in studying the properties of these dictionaries. 
Fig.~\ref{Fig:EncodingMKSVD_DBI} (top row) shows examples
of dictionaries from \cite{Mairal08},
which were trained on a large
database of natural noise-free images by a multiscale extension of the K-SVD algorithm \cite{Aharon06}. The scale is denoted by $s$ and the atom sizes are
indicated for each dictionary. The resulting encoding on the
Poincar\'{e} sphere for these dictionaries is also shown in
Fig.~\ref{Fig:EncodingMKSVD_DBI}. It can be seen that with increasing
the scale more directions become populated and the regularity
 of the atoms tends to increase. Indeed, if we look more closely in the left-most dictionary, corresponding to $s=0$, we can see that there are almost no diagonally oriented atoms, while there are a number of atoms with perfectly vertical and horizontal stripes. A number of noise-like and blob-like atoms are present, and these are responsible for the points near the center of the sphere. The dictionary in the middle (for $s=1$) already contains some diagonally oriented atoms, but still not as directional as it is the case for its many perfectly vertical and horizontal atoms. This is clearly visible in the corresponding Poincar\'{e} plot. The right-most dictionary indeed has atoms in various directions as it can be immediately concluded from the corresponding point cloud produced by our encoding tool.

\begin{figure*}[t]
\begin{center}
\begin{tabular}{cc}
\includegraphics[width=0.9\textwidth]{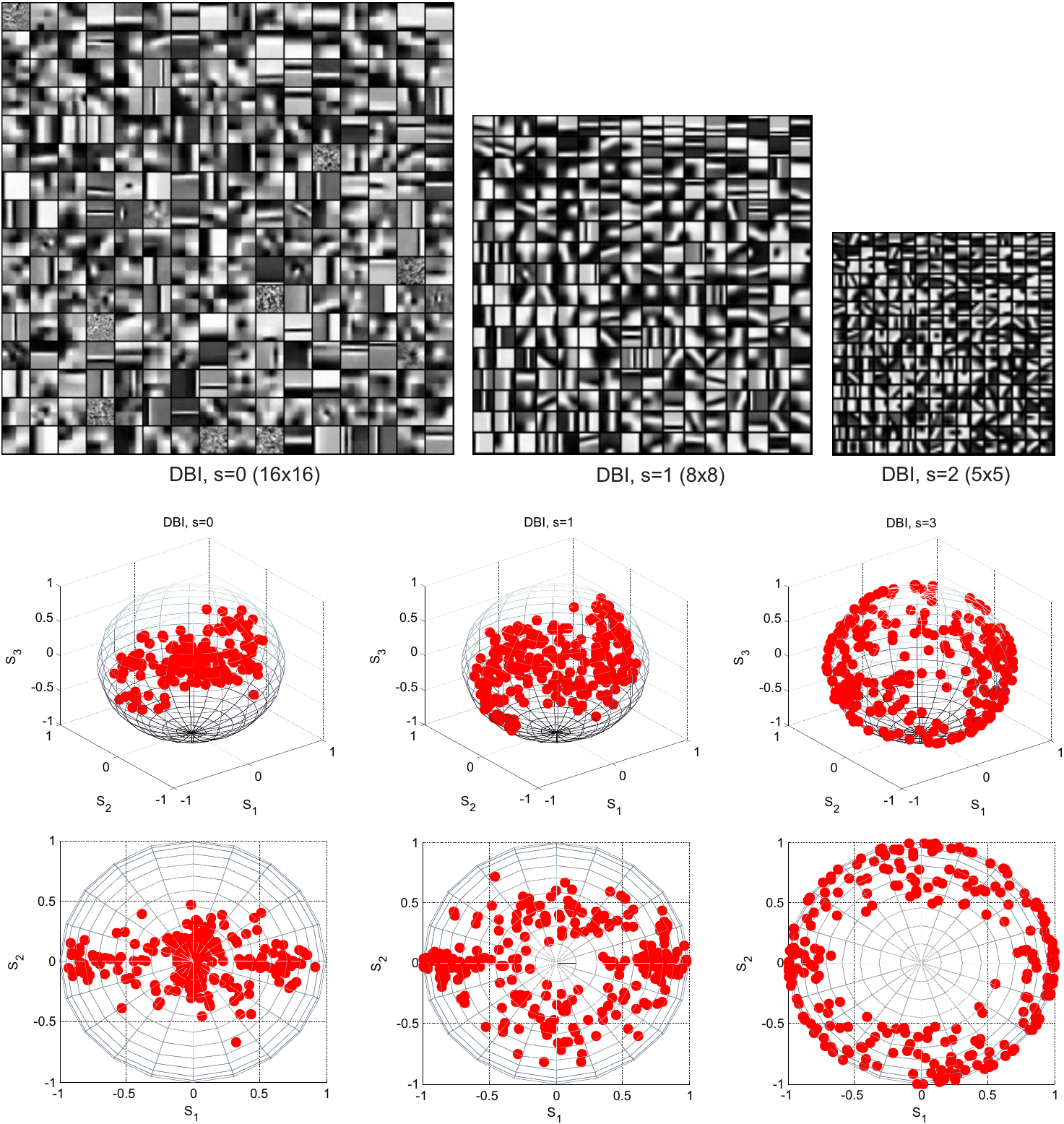}
\end{tabular}
\caption{Visualizing dictionaries of image atoms. {\bf Top:} a multiscale dictionary from
\cite{Mairal08} (Fig.~5),
which was trained over a large database of natural images. $s$ denotes the
scale and patch sizes are given in brackets. {\bf Bottom:} The corresponding Poincar\'{e} plots (using LDC-based regularity estimation). 
\label{Fig:EncodingMKSVD_DBI}}
\end{center}
\end{figure*}

\subsection{Generating dictionaries of image atoms}

So far we considered straightforward application scenarios of our encoding tool where we translate input image patches or atoms into point constellations on the Poincar\'{e} sphere. Now we take a reverse direction. Suppose that we have constructed an optimal (according to some criterion) point constellation or a spherical code, and we want to generate the corresponding dictionary of image atoms. Actually, this can be done quite simply, as we demonstrate next. All that we need to do is to read the three spherical coordinates ($\rho$, $2\psi$ and $2\chi$) of a point, or to calculate them from the Cartesian coordinates ($S_1$, $S_2$, $S_3$), and to generate a (random) patch that has the corresponding regularity, dominant direction and mean intensity.

Here we give an algorithm that generates random-bar atoms from point constellations. The main idea is to start from a binary patch with $L$ randomly placed horizontal lines (bars), where $L$ is determined from the elevation  $2\chi$, which is related to the normalized mean intensity $T$ via (\ref{Eq:NormalizedIntensity}). This pattern is then rotated to match the given azimuth $2\psi$ and finally, a randomization can be applied if $\rho<1$. With spherical codes, $\rho$ will be 1 for all the points. For constructing dictionaries from more general point constellations, with $\rho \leq 1$, the randomization can be applied by e.g. reshuffling (randomly) pixels until the specified $\rho$ is reached, or by adding noise. Finally each atom should be normalized, e.g. to a constant energy. A pseudo-code of this dictionary generation scheme is given in Algorithm \ref{Alg:Alg1}, which takes as an input an arbitrary 3D point constellation $\mathbf{S}$ and calls Algorithm \ref{Alg:Alg2} for generating a single random-bar atom with the corresponding attributes and the desired atom size.


Fig.~\ref{Fig:DictSphericalCodes} shows examples of random-bar dictionaries generated from spherical codes with 16, 72 and 256 points. The three point constellations were generated by an iterative algorithm of \cite{Lazic87}, described also in \cite{Pizurica98}, which places points on the sphere such that their minimal distance is maximized. In the same way, these random-bar dictionaries can be generated from arbitrary spherical codes \cite{Sloane81,Hardin96,ConwaySloaneBook} or from any arbitrary 3D point constellation.

Once the point constellation is available, generation of these dictionaries is extremely fast, for an arbitrary dictionary size and an arbitrary size of atoms. Although further research will be needed to explore the actual potentials of this dictionary design, initial results in image reconstruction experiments are quite encouraging. Fig.~\ref{Fig:ReconstructionsFromPD} illustrates reconstructions of a test image from a dictionary with 72 atoms of size $8 \times 8$ (shown in Fig.~\ref{Fig:DictSphericalCodes}, middle right) and from a dictionary with 256 atoms of the same size (shown in Fig.~\ref{Fig:DictSphericalCodes}, bottom). The reconstructions were performed using the orthogonal matching pursuit (OMP) algorithm \cite{Tropp07} implemented in the OMP-Box-v10 \cite{KSVDtoolbox}, where we selected the option with overlap and sparsity level 5. The quality of these reconstructions is actually comparable to those that are obtained with some dictionaries that are trained on natural images, as we demonstrate next.

We perfomed an experiment where a K-SVD dictionary \cite{Aharon06} was trained on one natural image (in particular, \emph{Barbara}) and then employed to reconstruct this image on which it was trained and four other images. We compared the resulting reconstruction performance with our random-bar dictionaries generated from a spherical code. In all experiments the atoms were of size $8 \times 8$; the K-SVD dictionary size was 256 and our random bar dictionaries were generated from a 256-spherical code (shown in Fig.~\ref{Fig:DictSphericalCodes}, bottom). We performed reconstructions with two random realizations of dictionaries from the 256-spherical code: $PD_{256,1}$ and $PD_{256,2}$ and with the union of these two random dictionaries: $PD_{2 \times 256}$. Peak-Signal-To-Noise-Ratio (PSNR) results are shown in Table~\ref{Tab:PSNR} and in Fig.~\ref{Fig:PSNRrec_complete}, where also the test images are shown. Several interesting observations can be made from this experiment: Firstly, two random-bar dictionaries generated from the same point constellation showed a very similar performance in the reconstructions (the difference in PSNR was less than 0.2dB in all cases and in most cases even much smaller). Secondly, the K-SVD dictionary was far superior over these random dictionaries on the image on which it was trained, but on other test images (that were not included in the training of the K-SVD dictionary), the random-bar dictionaries showed a very similar (and even slightly better) reconstruction results. Thirdly, using two random dictionaries improves, as expected, the reconstruction performance (in all cases for about 1dB). 

We repeated these image reconstruction experiments with a random-bar dictionary generated from an optimal covering with 1082 points on the sphere, with icosahedral symmetry \cite{SphericalCodesWebsite}, shown in Fig.~\ref{Fig:icover1082}. This dictionary, also shown in Fig.~\ref{Fig:icover1082} improves for all images the reconstruction PSNR for 0.6 - 0.8dB compared to the 512-dictionary $PD_{2 \times 256}$. 

\begin{algorithm}[t]
\caption{\emph{GenerateDictionary}($\mathbf{S}$, $N$)}
\label{Alg:Alg1}
\algsetup{indent=2em}
\begin{algorithmic}[1]
\STATE $NrAtoms=ReadLength(\mathbf{S})$  
\FORALL{$i=1:NrAtoms$}  
	\STATE $S_1=\mathbf{S}(i,1)$; $S_2=\mathbf{S}(i,2)$; $S_3=\mathbf{S}(i,3)$ 
	\STATE $\rho=\sqrt{S_1^2+S_2^2+S_3^2}$
	\STATE $\theta=\arcsin{(S_3/\rho)}$
	\STATE $\phi=\arcsin{(S_2/(\rho \cos(\theta)))} + [S_1<0]\pi$
	\STATE $Atom_i=GenerateAtom(\rho, \theta, \phi, N)$
\ENDFOR

\end{algorithmic}
\end{algorithm}

\begin{algorithm}[t]
\caption{\emph{GenerateAtom}($\rho$, $\theta$, $\phi$, $N$)}
\label{Alg:Alg2}
\algsetup{indent=2em}
\begin{algorithmic}[1]
\STATE $Atom=zeros(N,N)$;  \ \ \%initialize by a black patch
\STATE $T=\theta/\pi +0.5$
\STATE $L=round(TN)$
\STATE \textbf{Generate} a random index set $IND=[ind_1,...ind_L]$ such that $ind_i \neq ind_j$ for $i \neq j$, and $1 \leq ind_i \leq N$;
\FORALL{$i=1:L$}  
	\STATE $Atom(IND(i),1:N)=1$  \ \  \%place $L$ white horizontal lines
\ENDFOR
\STATE $Rotate(Atom, \phi)$ 
\IF{$\rho<1$}
\STATE $Randomize(Atom, \rho)$  \ \ \ \%reshuffling pixel intensities or adding noise
\ENDIF
\STATE $Normalize(Atom)$  \ \  \%normalize such that $\sum_i\sum_j Atom(i,j)^2=1$
\end{algorithmic}
\end{algorithm}

\begin{figure*}[t]
\begin{center}
\begin{tabular}{cc}
\includegraphics[width=\textwidth]{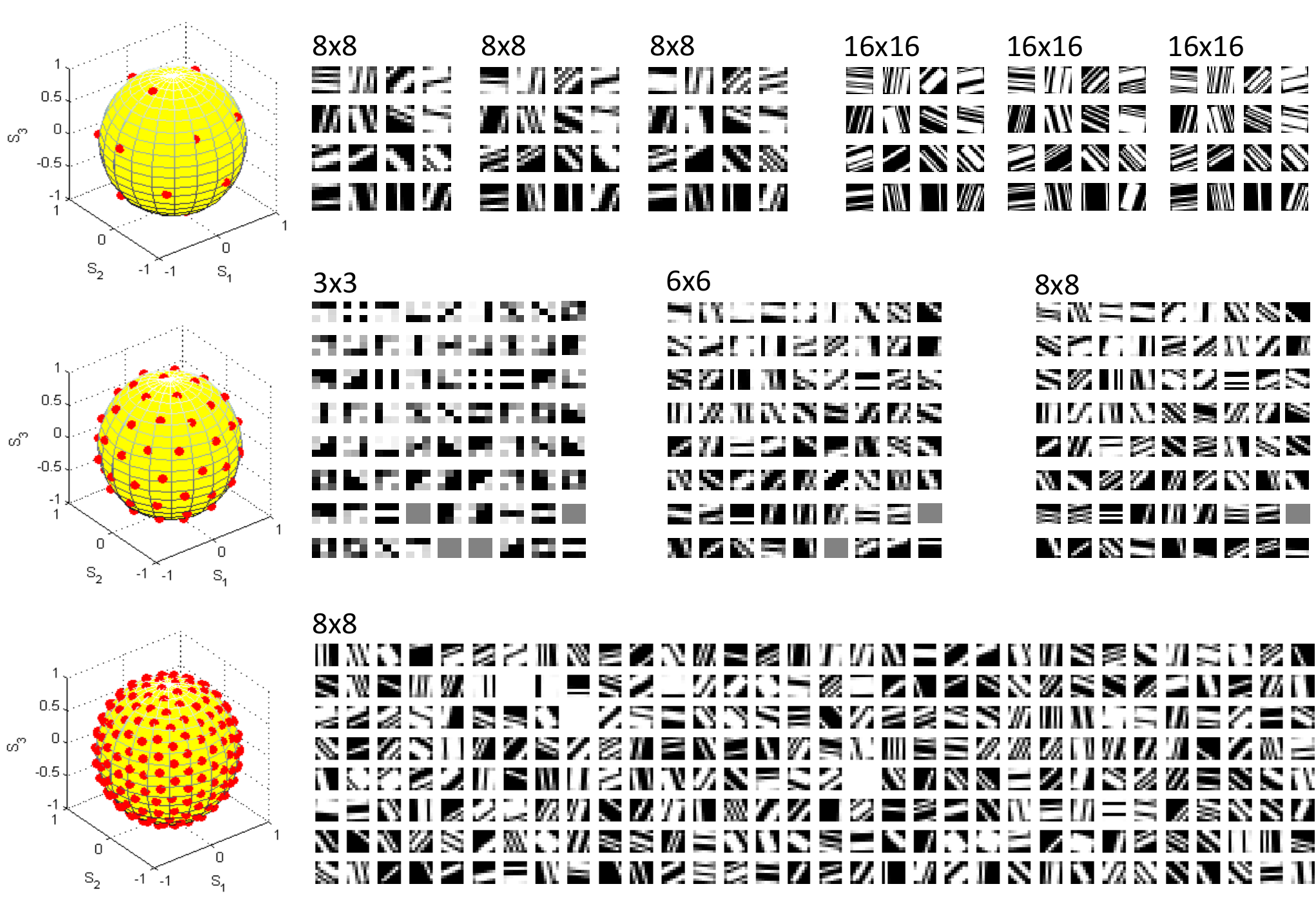}
\end{tabular}
\caption{Examples of random-bar dictionaries of image atoms generated from spherical codes. {\bf Top:} 16-points spherical code and generated dictionaries with $8 \times 8$ and $16 \times 16$ atoms (three realizations of each). {\bf Middle:} 72-points spherical code and examples of generated dictionaries with $3 \times 3$, $6 \times 6$ and $8\times 8$ atoms. {\bf Bottom:} 256-points spherical code and a generated dictionary with $8 \times 8$ atoms. 
\label{Fig:DictSphericalCodes}}
\end{center}
\end{figure*}

\begin{figure*}[t]
\begin{center}
\begin{tabular}{cc}
\includegraphics[width=\textwidth]{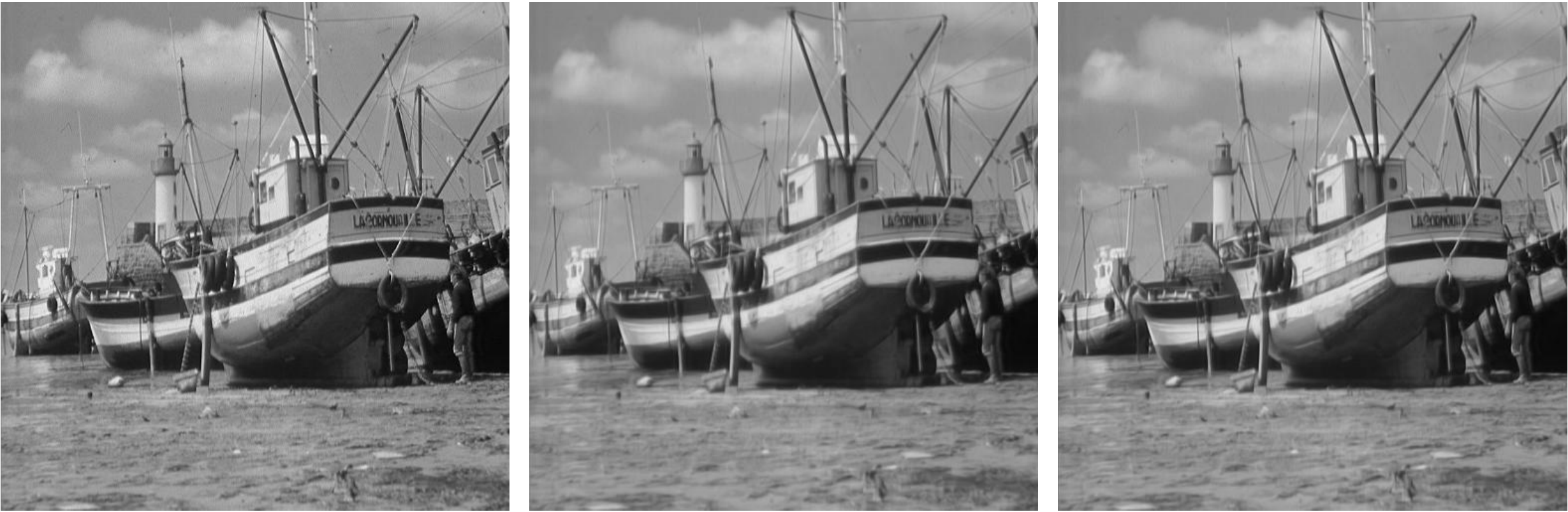}
\end{tabular}
\caption{An example of image reconstruction from random-bar dictionaries. Reconstruction method: OMP with overlap and sparsity level 5. {\bf Left:} original image. {\bf Middle:} Reconstructed from a dictionary with 72 atoms (in Fig.~\ref{Fig:DictSphericalCodes}, middle right). {\bf Right:} Reconstructed from a dictionary with 256 atoms (in Fig.~\ref{Fig:DictSphericalCodes}, bottom). 
\label{Fig:ReconstructionsFromPD}}
\end{center}
\end{figure*}

\begin{table}[t]\caption{PSNR[$\mathrm{dB}$] results of OMP reconstruction (with sparsity 5) using different dictionaries: \emph{K-SVD} with 256 atoms trained on \emph{Barbara} (marked by *), two random-bar dictionaries generated from the same 256-spherical code: $PD_{256,1}$ and $PD_{256,2}$, the union of these two dictionaries: $PD_{2 \times 256}$, and a dictionary generated from the 1082-point covering, $PD_{i1082}$. In all cases, atom size is $8 \times 8$.}\label{Tab:PSNR}
\begin{center}
\begin{tabular}{c||c|c|c|c|c}
\hline
 & \multicolumn{5}{c}{\scriptsize{Test image}}\\ \hline
 \scriptsize{Dict.} & \scriptsize{\emph{Barbara*}} & \scriptsize{\emph{Boat}} &
\scriptsize{\emph{House}} & \scriptsize{\emph{Peppers}} & \scriptsize{\emph{BrainMRI}}\\ \hline
 \scriptsize{\emph{K-SVD*$_{256}$}}
 & \scriptsize{34.40} & \scriptsize{33.38}&
 \scriptsize{36.38} & \scriptsize{35.36} & \scriptsize{30.15}
\\ \hline
 \scriptsize{$PD_{256,1}$}
 & \scriptsize{31.74} & \scriptsize{33.63}&
 \scriptsize{36.58} & \scriptsize{35.55} & \scriptsize{30.31} 
 \\ \hline
 \scriptsize{$PD_{256,2}$}
 & \scriptsize{31.83} & \scriptsize{33.47}&
 \scriptsize{36.51} & \scriptsize{35.41} & \scriptsize{30.16} 
  \\ \hline
 \scriptsize{$PD_{2 \times 256}$}
 & \scriptsize{32.65} & \scriptsize{34.33}&
 \scriptsize{37.60} & \scriptsize{36.16} & \scriptsize{31.04}
 \\ \hline
 \scriptsize{$PD_{i1082}$}
 & \scriptsize{33.40} & \scriptsize{34.95}&
 \scriptsize{38.43} & \scriptsize{36.77} & \scriptsize{31.69} \\
\hline
 \end{tabular}
 \end{center}
\end{table}

\begin{figure}[t]
\begin{center}
\begin{tabular}{cc}
\includegraphics[width=0.65\textwidth]{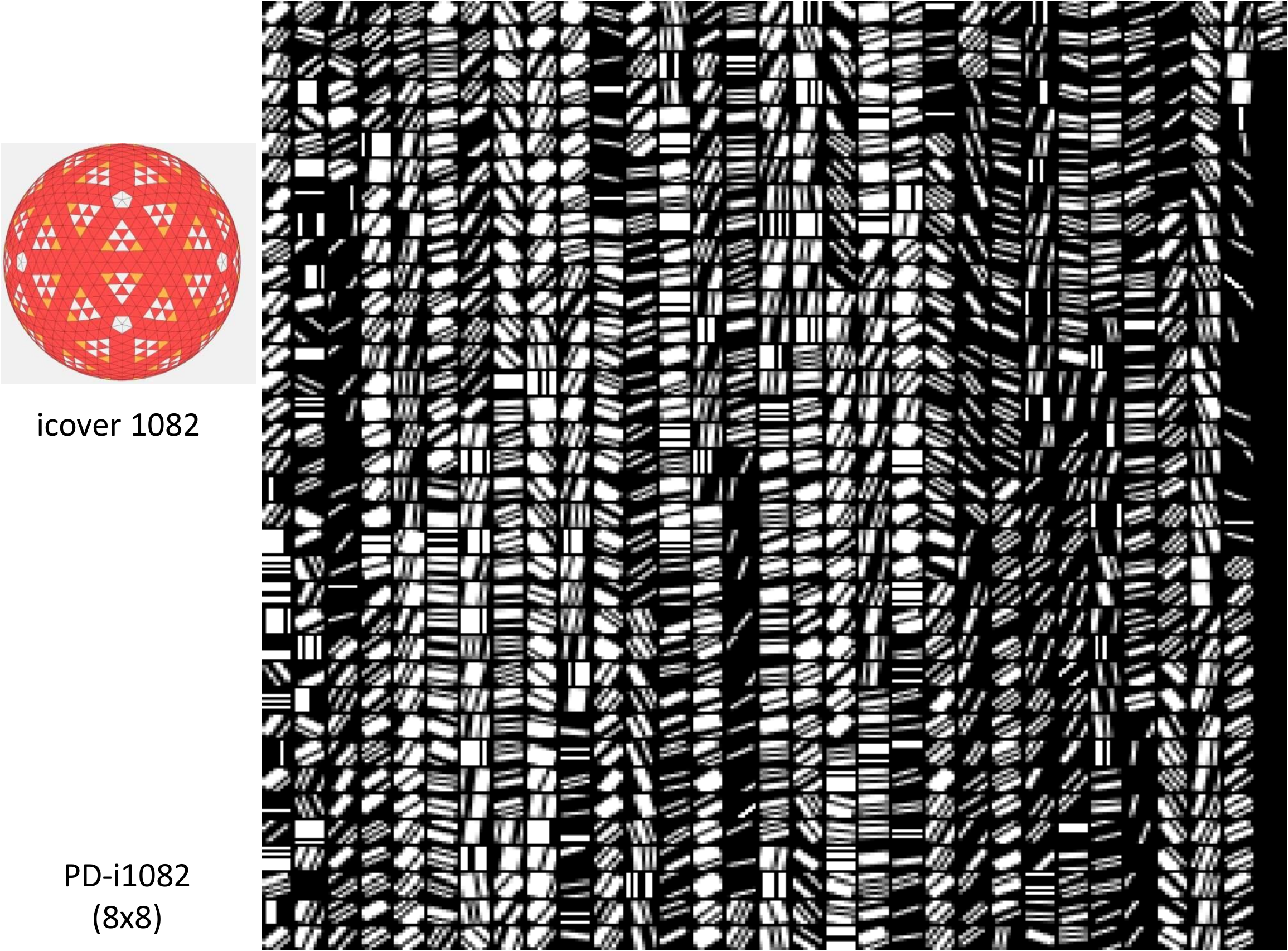}
\end{tabular}
\caption{Covering with 1082 points `icover1082' from \cite{SphericalCodesWebsite} (showing the convex hull of the points) and a random-bar dictionary $PD_{i1082}$ generated from this spherical code.
\label{Fig:icover1082}}
\end{center}
\end{figure}

\begin{figure}[t]
\begin{center}
\begin{tabular}{cc}
\includegraphics[width=0.65\textwidth]{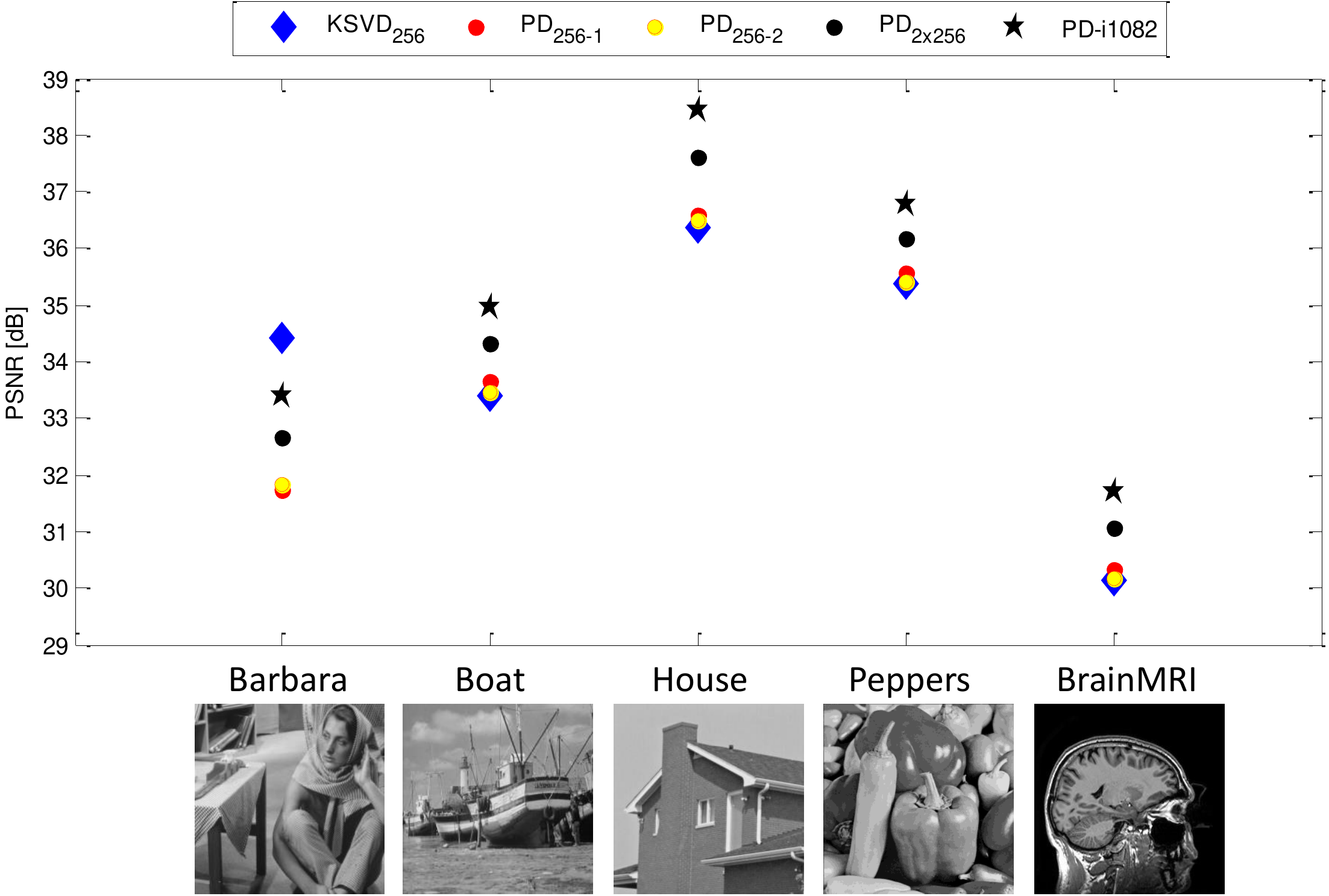}
\end{tabular}
\caption{Test images and PSNR results from image reconstruction experiments, listed in Table~\ref{Tab:PSNR}. The K-SVD dictionary was trained on \emph{Barbara}. 
\label{Fig:PSNRrec_complete}}
\end{center}
\end{figure}

\section{Conclusion} \label{Sec:Conclusion}

We introduced a graphical tool for encoding visual patterns, inspired by the Poincar\'{e} sphere in optics. A practical encoding scheme was developed within this framework, based on the following three features: dominant orientation, regularity and mean pattern intensity. Some possible generalizations were discussed, including extensions to four dimensions where the scale is also explicitly encoded. Three possible applications were demonstrated: patch clustering, visualizing properties of dictionaries of image atoms and generating random-bar dictionaries from spherical codes. Further work will focus on employing this tool in actual computer vision applications.

\section*{Acknowledgements}
I wish to thank Javier Portilla, especially for pointing to links with Wald features of texture perception and for inspiring private correspondence that initiated extensions and improvements of the initial version of the manuscript, and to Ingrid Daubechies for stimulating discussions that encouraged me to carry out this work.

\section*{Appendix}\label{Sec:AppendixProjectors}

\begin{figure}[t]
\begin{center}
\begin{tabular}{cc}
\includegraphics[width=0.55\textwidth]{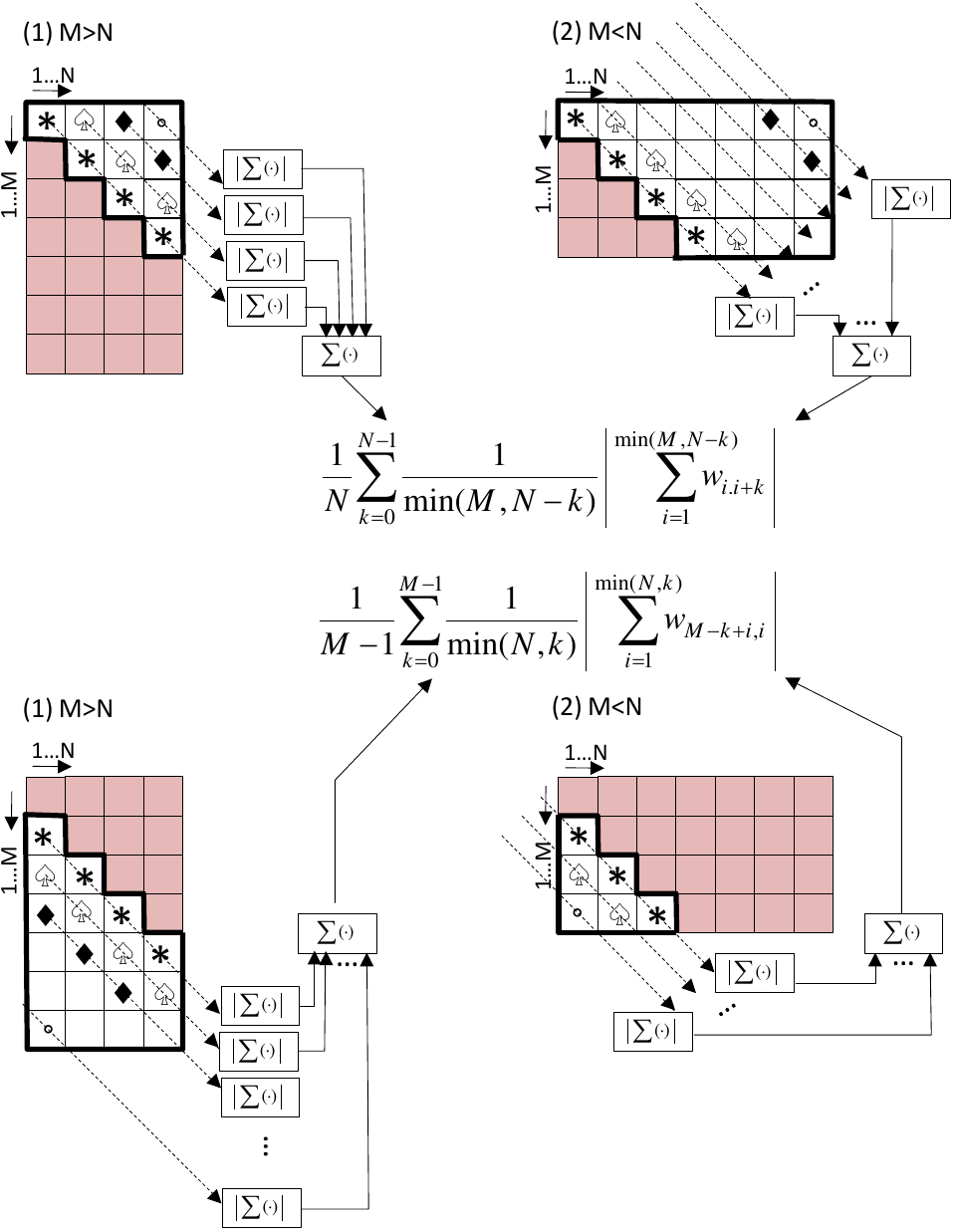}
\end{tabular}
\caption{Calculation of the first term (top) and the second term
(bottom) in the equation for $R_{d1}$.
\label{Fig:DiagonalProjectors}}
\end{center}
\end{figure}

Estimating dominant orientation of a zero mean patch $\mathbf{w}=\{w_{i,j}\}$ with the method of four Radon-like projectors from Section \ref{Sec:DominantOrientation} requires calculating
\begin{equation}\label{Eq:AppRk}
R_{k}=\frac{1}{\rho_k}\sum_{k=1}^{\rho_k}\Bigl|\frac{1}{|L_{k,l}|}\sum_{(i,j) \in L_{k,l}}w_{i,j}\Bigr|
\end{equation}
where $L_{k,l}$ denotes the set of points (projection ray) with the slope index $k$ and the
intercept index $l$, and $\rho_k$ the number of projection rays in the direction $k$, $k \in \{h,v,d_1,d_2\}$.

In horizontal and vertical directions, $L_{k,l}$ are simply rows and columns of the matrix, respectively, hence the expressions for $R_h$ and $R_v$ are obvious and were already given in Section~\ref{Sec:DominantOrientation}. The 135 degree projector, $R_{d1}$, for rectangular patches of size $M \times N$ can be expressed as:
\begin{equation}
R_{d1}=\frac{1}{N}\sum_{k=0}^{N-1}\frac{1}{\min(M,N-k)}\Bigl|\sum_{i=1}^{\min(M,N-k)}w_{i,i+k}\Bigr| 
+\frac{1}{M-1}\sum_{k=1}^{M-1}\frac{1}{\min(N,k)}\Bigl|\sum_{i=1}^{\min(N,k)}w_{M-k+i,i}\Bigr|
\end{equation}
where the first term results from projections and summation in the
upper right part (as shown in the top of
Fig.~\ref{Fig:DiagonalProjectors}) and the second from the same
operations in the lower left part (bottom of
Fig.~\ref{Fig:DiagonalProjectors}). In analogous way, the
projector for 45 degrees is constructed:

\begin{equation}
R_{d2}=\frac{1}{M}\sum_{k=1}^{M}\frac{1}{\min(N,k)}\Bigl|\sum_{j=1}^{\min(N,k)}w_{k-j+1,j}\Bigr|  
+\frac{1}{N-1}\sum_{k=2}^{N}\frac{1}{\min(N,M+k-1)-k+1} \\ \Bigl|\sum_{j=k}^{\min(N,M+k-1)}w_{N-j+k,j}\Bigr|
\end{equation}
where the first term results from summing the absolute values of
the projections in the upper left part of the image patch and the
second term results from the same operations in the lower right
part of the patch.

For square patches, these expressions simplify to:
\begin{equation}
R_{d1}=\frac{1}{M}\sum_{k=0}^{M-1}\frac{1}{M-k}\Bigl|\sum_{i=1}^{M-k}w_{i,i+k}\Bigr| 
+\frac{1}{M-1}\sum_{k=1}^{M-1}\frac{1}{k}\Bigl|\sum_{i=1}^{k}w_{M-k+i,i}\Bigr|
\end{equation}
and
\begin{equation}
R_{d2}=\frac{1}{M}\sum_{k=1}^{M}\frac{1}{k}\Bigl|\sum_{j=1}^{k}w_{k-j+1,j}\Bigr| 
+\frac{1}{M-1}\sum_{k=2}^{M}\frac{1}{M-k+1}\Bigl|\sum_{j=k}^{M}w_{M-j+k,j}\Bigr|
\end{equation}

\bibliographystyle{IEEEtran}


\end{document}